\documentclass[journal, compsoc]{IEEEtran}
\usepackage{amsmath,amsfonts}
\usepackage{algorithmic}
\usepackage{algorithm}
\usepackage{array}
\usepackage[caption=false,font=normalsize,labelfont=sf,textfont=sf]{subfig}
\usepackage{textcomp}
\usepackage{stfloats}
\usepackage{url}
\usepackage{verbatim}
\usepackage{graphicx}
\usepackage{cite}
\usepackage{multirow}
\usepackage{pifont}
\usepackage[inline]{enumitem}
\usepackage{hyperref}
\usepackage{xcolor}
\usepackage{soul}

\title{Introducing MeMo: A Multimodal Dataset for Memory Modelling in Multiparty Conversations} %with a focus on retention and encoding
\author{Maria Tsfasman, Bernd Dudzik, Kristian Fenech, Andras Lorincz,  Catholijn M. Jonker, Catharine Oertel}

\begin{document}

\IEEEtitleabstractindextext{%
\begin{abstract}
Conversational memory is the process by which humans encode, retain and retrieve verbal, non-verbal and contextual information from a conversation. Since human memory is selective, differing recollections of the same events can lead to misunderstandings and misalignments within a group. Yet, conversational facilitation systems, aimed at advancing the quality of group interactions, usually focus on tracking users' states within an individual session, ignoring what remains in each participant's memory after the interaction. Understanding conversational memory can be used as a source of information on the long-term development of social connections within a group. This paper introduces the MeMo corpus, the first conversational dataset annotated with participants' memory retention reports, aimed at facilitating computational modelling of human conversational memory. The MeMo corpus includes 31 hours of small-group discussions on Covid-19, repeated 3 times over the term of 2 weeks. It integrates validated behavioural and perceptual measures, audio, video, and multimodal annotations, offering a valuable resource for studying and modelling conversational memory and group dynamics. By introducing the MeMo corpus, analysing its validity, and demonstrating its usefulness for future research, this paper aims to pave the way for future research in conversational memory modelling for intelligent system development.
\end{abstract}

}

\maketitle
 \IEEEpeerreviewmaketitle
\IEEEdisplaynontitleabstractindextext

\section{Introduction}
\IEEEPARstart{H}{uman} memory for conversations plays a crucial role in shaping social bonds and fostering relationship building, as well as decision-making in future interactions \cite{Bluck2005FunctionsMemory}. Understanding human conversational memory is, thus, essential for explaining and predicting human behaviour in conversations. Conversational memory can be defined as a sub-type of episodic memory, which manages the encoding, storage, and retrieval of personally experienced events \cite{NeuroEncyclopedia2021}, particularly within conversational settings. Previous research on conversational memory shows that numerous factors can affect what is encoded and retained from a conversation: the relationship between participants, their characteristics, linguistic features of produced speech and many more \cite{mckinley_memory_2017, benoit_participants_1996, Samp2007_FriendsMemory, Knutsen2014EgoMemory, Miller_Interpersonal_2002, diachek2024LinguisticFeaturesPredictRecall}. However, due to the multitude of variables involved, it remains unclear how these factors interact to determine which memories are more likely to be retained and which are more likely to be forgotten over time.
      
One way of investigating the intricate potentially non-linear relationship between contributing factors to a socio-cognitive process such as memory is by creating a computational model. Although never applied to conversational memory, to our knowledge, other socio-cognitive phenomena such as affect, engagement or cohesion have been previously investigated using computational models \cite{poria2017AffectiveComputingReview, Vinciarelli2009SSP, cambria2014NLP}. Researchers model these socio-cognitive processes by training machine learning algorithms to predict user's internal states from verbal and non-verbal data. A major issue to take into account when building such a model is the data used for its creation since the model can only be as accurate as the data it is trained on \cite{Bender2021Parrots}. Similar to the reproducibility crisis in the field of social sciences \cite{nosek2022replicability}, there is more and more understanding of how a biased or misconstrued dataset can be detrimental to the reproducibility, generalisability and validity of the resulting models \cite{hutson2018AIreproducibility, Paullada2021MLEcologicalVal}. Many scholars are calling for more careful creation of datasets aimed at computational modelling \cite{hutson2018AIreproducibility, Paullada2021MLEcologicalVal,  Bender2021Parrots}. Therefore, constructing a computational model of conversational memory requires a dataset that is representative of the modelled constructs and is collected in an ecologically valid setting. 

%Inroducing MeMo corpus
Since there is no such data available for conversational memory research, in this paper, we introduce the \textbf{MeMo} (\textbf{Me}mory \textbf{Mo}delling) corpus - the first conversational corpus with annotations for conversational memory. The \emph{MeMo} corpus is aimed to be used for computational modelling of conversational memory developed with multidisciplinary research in mind. It, therefore, combines validated behavioural and perceptual measures as well as video, audio, and multimodal annotations, individual and group eye gaze behaviour, head pose, low-level hand gestures, and text. The variety of measures, multimodality and ecological validity of the corpus make it a useful resource for computational as well as behavioural studies on conversational memory and group dynamics. Because of the data complexity, the dataset will be released in batches with the process described in \emph{Section \ref{data_release}} once the pseudo-anonymisation process is complete. In this paper, we aim to achieve the following goals:
\begin{itemize}
\item \textbf{Introducing the \emph{MeMo} corpus.} We describe the \emph{MeMo} corpus, the data collection, and its challenges. The \emph{MeMo} corpus pioneers a way of collecting first-party memory annotations directly usable in computational research - by combining a free-recall task with a subsequent first-party annotation of the recorded memorable moments to a video time frame (see \emph{Section \ref{procedure}}).

\item \textbf{Demonstrating its usefulness.} We demonstrate how the corpus can be used to build conversational memory models and summarise empirical results on the corpus (see \emph{Section \ref{ML}}).

\item \textbf{Suggesting potential topics of future research using the corpus.} We describe potential modelling tasks that can be explored with the use of the \emph{MeMo} corpus (see \emph{Section \ref{uses}}).
\end{itemize}

\section{Motivation for creating MeMo}
\label{motivation}

%In this section, we provide a detailed motivation for creating the MeMo dataset and introduce the main goals the dataset aims to achieve.
Humans are inherently social creatures, and the quality of one's social connections significantly impacts their psychological and physiological well-being \cite{Holt‐Lunstad2021SocialConnectionsAndHealth}. Feeling listened to, understood, and appreciated within a social relationship is essential for fostering these quality connections \cite{Reis2000Relatedness}. In group settings, such as family gatherings or work meetings, it could be challenging to achieve this quality, because of differences in personality, dominance and other factors. 
Conversation facilitation has emerged as a promising approach to enhancing the quality of these group interactions \cite{Garcia1991Mediation, Picard2011DeepeningConversations}. Over repeated sessions with a trained facilitator, groups of people can resolve conflicts, deepen conversations, and foster mutual understanding across various social settings \cite{Dillard2013DeliberationFacilitation, Garon2002FacilitatingMeetings, Bruce2017FamilyMeetingSupport}. 
Conversation facilitation can be challenging, requiring undivided attention towards multiple team members, conversation structure and content \cite{Hayne1999GroupSupportFacilitation}. Computational systems can support human facilitators in this task, as well as serve as an alternative solution in the absence of a human facilitator \cite{Phillips1993Faciliated}.

Existing conversation support systems have been shown to improve social interaction satisfaction, encourage equal participation and decrease social inhibitions \cite{Lindblom2009LargeMeetingSupport, Shamekhi2019RealtimeMeetingSupport, Li2022VRTurntakingSupport}. A common method of achieving these results involves continuous tracking of users' non-verbal and verbal signals \cite{Schiavo2014OvertGroupSupport}. These low-level signals are then used to infer a real-time measure of users’ participation (e.g.\cite{Shamekhi2019RealtimeMeetingSupport}) or more complex internal states such as attention (e.g. \cite{Kulyk2005MeetingSupportAttention}), dominance (e.g. \cite{Kim2008MeetingSupportDominance}) or social presence (e.g. \cite{Nowak2023MeetingSupportSocialPr}). Based on these predicted measures, a system then produces suggestions on how to enforce meeting structure and promote equal participation (e.g. \cite{Schiavo2014OvertGroupSupport}, \cite{Shamekhi2019RealtimeMeetingSupport}).
These predicted measures are real-time indicators of a participant's immediate reactions or internal states, captured at specific timestamps or as cumulative measures from the start of the session. Although these measures can be used to represent users' current state or trends within a session, they do not always represent the way the user will feel about the subject in subsequent interactions, since feelings triggered by an event within an interaction could be forgotten or completely changed over time, in retrospect to the event \cite{Okada2019Affective}. To sum up, user experience itself may not matter as much as the user’s memory of that experience in the context of long-term interaction and future decisions \cite{Norman2009MemoryVSExperience}.  Consequently, for long-term interaction, a facilitation system needs to not only track the current state of the user but also understand the user's memory of conversational experiences.

Decades of cognitive research show that, due to the selective nature of human memory, only a fraction of perceived experiences are encoded and retained \cite{wixted_memory_2018}. The (subconscious) decision to retain or forget a conversational event is not random, it is affected by an intricate combination of inter- and intra-personal factors, such as conversational context, linguistic parameters of speech, one's role in the conversation, and conversational skills \cite{Knutsen2014EgoMemory, Miller_Interpersonal_2002, diachek2024LinguisticFeaturesPredictRecall}. Apart from these over-arching factors, according to previous research, humans tend to retain experiences that (1) enrich or confirm their self-image \cite{Knutsen2014EgoMemory}, (2) connect them to other individuals \cite{Morris_Conversation1993O, Miller_Interpersonal_2002, Bietti2010SharingMF} or (3) guide their future actions, thoughts and responses \cite{Bluck2005FunctionsMemory}.
While previous research has identified these factors and functions of human memory, how they operate together in spontaneous settings, determining whether an event will be retained or forgotten remains unclear.
Corresponding to the mentioned memory functions, understanding what remains in a user's memory after an interaction could help a facilitation system in (1) understanding a user's personal preferences and identity, (2) keeping track of relational development between conversational partners and (3) understanding the origin of perspectives and decisions in future interactions. Yet, to our knowledge, users' memory of the interaction has never been collected, analysed or inferred to advance meeting support systems.

Unlocking the potential of personal memories for user-modelling and personalisation purposes in conversational settings requires predictive models based on real-world conversational data. In another context - that of media consumption - researchers have built such models to predict which images or videos are more likely to be encoded and retained by a human viewer based on media features and user characteristics \cite{MemorabilityModeling2020, videoMemorability2018, Isola2014ImageMemorability, Bainbridge2017MemorabilityNeural}.
However, the conversational context is different from media consumption: unlike media consumption, conversational context involves continuous production and comprehension of verbal and non-verbal signals, involving different cognitive mechanisms \cite{stafford_actor-observer_1989} and producing qualitatively and quantitatively different memories \cite{benoit_participants_1996}. Moreover, conversational memory has various context-specific factors at play that do not apply to media consumption tasks: conversation-specific verbal and non-verbal signals, the relational dynamics between conversational partners, and many more \cite{mckinley_memory_2017, Samp2007_FriendsMemory, Knutsen2014EgoMemory, Miller_Interpersonal_2002, diachek2024LinguisticFeaturesPredictRecall}.
Therefore, the task of modelling how humans encode, retain and retrieve conversations remains unsolved. 

Since there has not been any computational research on the topic and the data used in the psychological research on conversational memory is mainly collected from controlled experiments, an essential step towards conversational memory prediction is constructing a dataset of spontaneous conversations annotated with memory reports. Therefore, we constructed the \textbf{\emph{MeMo} corpus to provide a resource for the two primary goals}: \textbf{(G1)} for research and computational prediction of participants' memory in spontaneous conversations via verbal and non-verbal signals and \textbf{(G2)} for the creation of conversational memory models supporting meeting facilitation in the context of repeated interactions.

\section{Guiding principles for collecting MeMo}
\label{principles}
When constructing a dataset suitable for studying and modelling human conversational memory, we argue that several major principles need to be considered. %In this section, we summarise the principles that guided the design of MeMo dataset and refer to them in \emph{Section \ref{procedure}} in relation to particular methodological decisions.

\subsection{P1: Maximising ecological validity}
While scraping the internet for datasets has become very common in computer science, researchers increasingly advocate for carefully curated datasets to better predict human behaviour in natural settings \cite{Bender2021Parrots, Heale2015Validity, Paullada2021MLEcologicalVal}. This is particularly important when the computational models are aimed to predict and explain human behaviour in a setting with minimal structural constraints, such as free-flowing conversations \cite{Heale2015Validity}. In the context of conversational memory modelling, this is particularly important since an unnatural, scripted setting can change the structure and the content of memories \cite{Rumpf2019EcolValMemory, Dunsmore2022SalientAttractMemory}. Therefore, we strove to ensure that the \emph{MeMo} corpus accurately reflects the conditions and variables present in real-life conversations.

\textbf{P1.1 Preserving Natural Interaction Environment.} First, the dataset would need to be recorded in a natural environment rather than in a laboratory setting. The laboratory environment can introduce artificial constraints and biases that may not exist in real-life conversational settings, including intrusive sensors and unnatural conversation settings (e.g. a lab with visible sensors and no natural light). It has been shown, that people perform differently in memory tasks in a laboratory setting in comparison to a real-world setting \cite{Schnitzspahn2018InLabMemory}. The dataset, therefore, needs to preserve a natural conversational setting typical to real-world environments.

\textbf{P1.2 Preserving Spontaneity of Conversation Interactions.} Second, for conversational memory reports to represent the processes engaged in an in-the-wild conversation or meeting, the conversation must be as spontaneous as possible. The main reason is that the processes involved in comprehension and production of spontaneous speech are different from reading out text or following a script (as in scripted corpora, e.g. \cite{Jiang_Zhang_Choi_2020scripted}). For example, memory for self-produced statements can differ from reading or hearing a statement \cite{Knutsen2014EgoMemory, Daneman1986Individual, stafford_actor-observer_1989}. Letting the conversation flow emerge by itself rather than imposing lab-created tasks or structure as much as feasible is important for the ecological validity of such conversational data.

\textbf{P1.3 Ensuring Representativeness of Participants.} Lastly, it is important for a dataset to recruit a representative sample of participants from diverse demographics and backgrounds. It is, unfortunately, a common practice in datasets and experimental studies to mainly recruit university students and staff, biasing the data towards a demographic of highly educated English-speaking white young women \cite{Erba2021CommunicationResearchDemographicsBias, Infante-Rivard2018SelectionBias}. An alternative to university students is a more diverse demographic of participants recruited through specialised websites such as Amazon Mechanical Turk. While this method provides a more diverse demographic, participants recruited this way might be 'professional participants' who go through a multitude of studies daily and are therefore biased in how they respond to the experiment questions \cite{ Qureshi2022AmazonTurkDemographics}. In either case, study results can greatly depend on the demographics of its participants \cite{Cohendet2016MemorabilityByGender}, and it is, therefore, important to report the demographics along with the dataset to understand the limitations of the data. So far, unfortunately, it is not a common practice and many datasets do not report the demographics of their participants (see section \emph{Section \ref{diversity}} for examples).

\subsection{P2: Maximising the construct validity of conversational memory measure} 
A principal goal of developing MEMO is to collect a corpus that facilitates the identification of moments in a conversation likely to be retained by its participants (\emph{$\rightarrow$G2}). Collecting such data is challenging due to the fundamentally different organisation of human experience as context-delineated episodes and the typically timestamp-delineated segments used to annotate moments in multimodal data (see Dudzik et al. \cite{Dudzik2018Lifelog} for a discussion). Aiming for construct validity involves ensuring that our chosen measures accurately reflect conversational memory while also recognising the limitations of the selected metric.

The existing approach to estimating which events are encoded from a conversation is a free-recall task - asking participants to report what they remember from the conversation in their own words, usually in writing \cite{Cleary2018MemoryMeasures}. To then access the events that these reports refer to for analysis, external annotators review the reports and identify the events mentioned within the conversation \cite{benoit_memory_1990, diachek2024LinguisticFeaturesPredictRecall}. Outsourcing this task to external observers may impact the construct validity of the resulting memory measure, as multiple moments might match a description, making it difficult to determine the specific event without asking the participant directly.

Given that the \emph{MeMo} corpus is designed to model conversational memory, we propose an alternative method. After a conversation, participants complete a free-recall task, describing what they remember. They then watch a recording and pinpoint the exact moments that match their memories (see \emph{Section \ref{sec:memory_annotation}} for details). This approach ensures that the identified moments accurately reflect the participants' memories, preserving the validity of the memory annotations. These annotations provide a reliable temporal link between memory reports and conversation segments, serving as ground-truth labels for computational modelling. 

\subsection{P3: Considering Context-sensitivity of Memory Processes}

Maximising ecological validity (P1) might imply that fewer variables are controlled (e.g. more diverse demographic, an in-the-wild experimental setting, etc.). Therefore, while maximising the internal validity of the data, it might reduce the external validity \cite{Patino_Ferreira_2018ValidityTradeoff}.
To avoid this, it is particularly important to track as many potential confounding variables as possible using validated questionnaires accepted by the scientific community. 

Specifically, conversational memory can be affected by communication skills \cite{Miller_Interpersonal_2002}, mood  \cite{Jd1983MoodRecall, Matt1992Mood-congruent}, personality \cite{Mayo1989PersonalityMemory}, values \cite{Villaseñor2021ValuesMemory} and the relationship dynamics between participants \cite{Samp2007_FriendsMemory, IOS_woosnam2010inclusion}. In addition, factors concerning group perception should be measured such as group entitativity, cohesion and rapport. While they have not been investigated in relation to conversational memory the research shows that they can influence learning \cite{Evans1991CohesionPerformance, Kim2020CohesionLearning}, which is inherently related to memory.

\section{Related work}
To the best of our knowledge, there are no datasets focusing on conversational memory. However, to contextualise \emph{MeMo}, we describe existing datasets that are aimed to support computational modelling research on memory in one way or another. In addition, we describe most related behavioural studies on the topic of conversational memory. We compare the data designs using the criteria most relevant for the context of the \emph{MeMo} corpus as shown in Table \ref{fig:related_datasets}. We describe each criterion in the following subsections.

\begin{table*} 
\centering
\label{fig:related_datasets}
\caption{The comparison between MEMO and related corpora \cite{cohendet_videomem_2019, Bernd_mementos, WoNoWA2020, Shaw2007homebirth} and the most related study \cite{diachek2024LinguisticFeaturesPredictRecall}
\textit{(Part. count - number of participants)}) }
\begin{tabular}{cccccccccc}

\hline\hline
\textbf{\begin{tabular}[c]{@{}c@{}}Dataset/\\ Study id\end{tabular}} & \textbf{\begin{tabular}[c]{@{}c@{}}Memory\\ sub-process\end{tabular}} & \textbf{\begin{tabular}[c]{@{}c@{}}Memory\\ task\end{tabular}} & \textbf{\begin{tabular}[c]{@{}c@{}}Recorded\\ behaviour\end{tabular}} & \textbf{\begin{tabular}[c]{@{}c@{}}Perceptual\\ measures\end{tabular}} & \textbf{\begin{tabular}[c]{@{}c@{}}Task\\ context\end{tabular}} & \textbf{\begin{tabular}[c]{@{}c@{}}Part.\\ count\end{tabular}}  & \textbf{\begin{tabular}[c]{@{}c@{}}Time\\ (h)\end{tabular}}  & \textbf{\begin{tabular}[c]{@{}c@{}}Longi-\\ tudinal\end{tabular}} \\ \hline \hline
\textbf{\begin{tabular}[c]{@{}c@{}}Video\\ Mem\end{tabular}} & \begin{tabular}[c]{@{}c@{}}encoding \\ \& retention\end{tabular} & \begin{tabular}[c]{@{}c@{}}recognition \\ (short \& \\ long-term)\end{tabular} & - & - & \begin{tabular}[c]{@{}c@{}}media\\ consumption\end{tabular} & 3246 & 19.4 & \ding{51} \\ \hline
\textbf{EEGMem} & encoding & \begin{tabular}[c]{@{}c@{}}recognition\\ (long-term)\end{tabular} & \begin{tabular}[c]{@{}c@{}}EEG\\ recording\end{tabular} & - & \begin{tabular}[c]{@{}c@{}}media\\ consumption\end{tabular} & 12 & 8.3 & -  \\ \hline
\textbf{Mementos} & retrieval & \begin{tabular}[c]{@{}c@{}}auto-\\ biographic\\ retrieval\end{tabular} & video & \begin{tabular}[c]{@{}c@{}}personality,  \\ mood, affect\end{tabular} & \begin{tabular}[c]{@{}c@{}}media\\ consumption\end{tabular} & 300 & 33 & - \\ \hline
\textbf{WoNoWa} & \begin{tabular}[c]{@{}c@{}}retrieval \\ \& use in \\ collabora-\\ tion\end{tabular} & \begin{tabular}[c]{@{}c@{}}transactive\\ memory\\ perception\end{tabular} & \begin{tabular}[c]{@{}c@{}}video, audio, \\ transcripts\end{tabular} & \begin{tabular}[c]{@{}c@{}}perceived \\ leadership \\ \& group \\ performance\end{tabular} & \begin{tabular}[c]{@{}c@{}}collaboration\\ task (group)\end{tabular} & 45 & 17 & -  \\ \hline
\textbf{\begin{tabular}[c]{@{}c@{}}Home Birth\\ helpline\end{tabular}} & retrieval & \begin{tabular}[c]{@{}c@{}}spontaneous\\ retrieval\\ in dialogue\end{tabular} & transcripts & - & \begin{tabular}[c]{@{}c@{}}spontaneous\\ conversation\\ (dyad)\end{tabular} & 56 & \begin{tabular}[c]{@{}c@{}}NA \\ (80\\ calls)\end{tabular} & \ding{51}\\ \hline
\textbf{\begin{tabular}[c]{@{}c@{}}Diachek \\ et al. 2024\end{tabular}} & \begin{tabular}[c]{@{}c@{}}encoding \\ \& retention\end{tabular} & \begin{tabular}[c]{@{}c@{}}free recall\\  reports\end{tabular} & \begin{tabular}[c]{@{}c@{}}disfluencies \\ from\\ transcripts\end{tabular} & - & \begin{tabular}[c]{@{}c@{}}spontaneous\\ conversation \\ (dyad)\end{tabular} & 118 & 14.8 & -\\ \hline \hline
\textbf{MEMO} & \begin{tabular}[c]{@{}c@{}} encoding\\ \& 
retention\end{tabular} & \begin{tabular}[c]{@{}c@{}}free recall \\ reports +\\ timing \\ annotation\end{tabular} & \begin{tabular}[c]{@{}c@{}}video, audio, \\ transcripts\end{tabular} & \begin{tabular}[c]{@{}c@{}}individual,\\ task, group\\ \& others'\\ perception\end{tabular} & \begin{tabular}[c]{@{}c@{}}spontaneous\\ conversation \\ (group)\end{tabular} & 53 & 31 & \ding{51} \\ \hline\hline
\end{tabular}
\end{table*}

\subsection{Facilitated Modelling Perspective}

When it comes to datasets for modelling memory processes, we distinguish between two different modelling perspectives that they facilitate: \begin{enumerate*}
    \item Corpora supporting \textit{Situation-centered} perspectives facilitate modelling how specific properties of a defined situation (e.g., exposure to a video) are expected to give rise to memory responses in members of some population (e.g., how specific video content is likely to be remembered by people in general). Datasets focusing on a situation-centred modelling perspective often attempt to capture a large range of distinct situations but typically have a small number of distinct individuals responding to them (e.g. \cite{cohendet_videomem_2019}).   
    In contrast, datasets with an \item \textit{Individual-centered} perspective typically focus on more fine-grained modelling of variation in memory processes across specific individuals, possibly considering interactions with the situation (e.g., ways in which individuals behaviorally express when specific video content triggers a memory in them \cite{Bernd_mementos}). Datasets focusing on supporting an Individual-centered modelling perspective often contain only a relatively small number of distinct situations but a relatively large number of individuals responding to them.   
\end{enumerate*} Note that datasets facilitating an Individual-centered perspective can often also support a Situation-centred one, but not the other way around (because responses are typically aggregated from individual responses to the situation level).

In \emph{MeMo}, we aim to combine the two perspectives as much as the conversational context permits. From an individual-centred perspective, \emph{MeMo} includes various perceptual and audio-visual measures from individual participants, providing resources to study how humans behave during memorable moments. From a situation-centered perspective, the data design allows for the investigation of the entire situation using audio-visual data from all group members, along with memory data aggregated across the group. This approach helps identify what makes a moment more memorable for a set of participants, abstracting from individual differences (for an example see \emph{Section \ref{uses}}).

\subsection{Memory process}
%memory process
Memory-related datasets vary in their primary goal, specifically the memory sub-process they aim to investigate. Human memory can be divided into three sub-processes: \emph{memory encoding} (processing the experience), \emph{memory retention} (preserving the experience), and \emph{memory retrieval} (extracting the retained experience) \cite{wixted_memory_2018}. 

These processes are closely intertwined with each other. For example, any study of memory involves some measure of memory retention - whether or not a memory was preserved and for how long. Whether a moment has been retained or not cannot be completely measured, since some memories might have been retained but are not accessible at the moment of the measure \cite{Tulving1966AvailabileAccessibleMem}. Therefore, when it comes to retention, researchers usually focus on investigating memories available for retrieval at the moment of a memory test. Forgetting then refers to the moments that are not available at the moment of collecting the memory measure.  Studies that focus on the retention process usually investigate the forgetting curve - collecting memory reports at several points in time and seeing how much information will be retained across the time \cite{Murre2015ForgettingCurve}. Two datasets have collected such data in the context of media retention and forgetting \cite{videoMemorability2018, 2022MedievalEEGMem}. To our knowledge, there have been no studies investigating retention or forgetting in free-flowing conversations.

Most measures of memory involve memory retrieval - for example, free-recall tasks, that ask participants to freely report what they recall from the given stimulus \cite{Cleary2018MemoryMeasures}. This said, most of these studies use retrieval as a memory-measuring tool to study encoding and retention. In contrast, memory retrieval dataset papers investigate the moments when memories are (spontaneously) triggered and extracted - whether it is a memory from childhood prompted by music videos \cite{Bernd_mementos} or memories relevant to a collaboration task at hand \cite{WoNoWA2020}. In conversational context, to our knowledge, only one dataset approaches moments of retrieval: \cite{Shaw2007homebirth} investigate how callers and call takers indicate to each other that they are speaking for a second or subsequent time, therefore retrieving memories of past interactions. 

While studies of retrieval focus on the moment a memory is being extracted, memory encoding investigation focuses on the specific stimulus or human response to the stimulus at the very moment the memory is being encoded. This means, that while the memory task itself might involve retention and retrieval (e.g. free-recall or recognition \cite{Cleary2018MemoryMeasures}), the focus of the study is the specific stimulus viewed or behaviour displayed during the event mentioned in the reported memory.
For example,  \cite{MemorabilityModeling2020, cohendet_videomem_2019, videoMemorability2018} that investigate the features of memorable media or \cite{2022MedievalEEGMem} investigating the brain signals at the moment of viewing the media that is to be retained. In conversational context, \cite{diachek2024LinguisticFeaturesPredictRecall} has previously investigated how linguistic features predict whether or not a conversational event will be encoded.

The \emph{MeMo} dataset has been designed to study and model two sub-processes of human episodic memory. First, the corpus allows for studying when information has been \emph{encoded} through the investigation of first-party temporal labels of events registered in participants' free-recall reports (see \emph{Section \ref{sec:memory_annotation}}). Second, the corpus is designed to study conversational \emph{retention}, with memory reports collected immediately after interactions for short-term memory and after 3-4 days for longer-term memory (see more detail in \emph{Section \ref{sec:memory_annotation}}; also notice the limitations for this task in \emph{Section \ref{sec:limitations}}). In principle, \emph{MeMo} could also be used for modelling memory retrieval processing during conversations (e.g., in terms of how people use memories for social bonding purposes during interactions \cite{Bluck2005FunctionsMemory}). However, the annotations provided with the current release do explicitly support this task (see \emph{Section \ref{sec:limitations}} for a discussion).

\subsection{Recorded behaviour \& measures}
Some datasets for memory research involve recording \textbf{participants' behaviour} and/or \textbf{perceptual measures} (i.e. self-reported individual traits or questionnaires on participants' perception of the task and other participants). These measures can vary from EEG brain signal recordings during the task \cite{2022MedievalEEGMem} to video of participants' nonverbal signals throughout the task (in \cite{Bernd_mementos}). In datasets involving conversation or interaction between participants, typically, there is also a recording of the speech, in the form of audio or transcripts \cite{WoNoWA2020, Shaw2007homebirth}. In addition to recorded behaviour, some datasets measure various self-reported perceptual measures \cite{Bernd_mementos, WoNoWA2020}. Since the \emph{MeMo} dataset is focused on human behaviour and perception in the conversational context, it includes video and audio recordings for behavioural measures and various self-reported measures of participants' individual characteristics, their perception of the interaction, group, and other participants. 

\subsection{Task context \& memory task}
An important parameter informing corpus design is the context of the task participants perform, such as solitary media consumption or human-human interaction. This context is closely linked with the memory reporting measure. For media consumption, memory is often measured with recognition tasks where participants identify previously seen videos, indicating memory retention or forgetting \cite{cohendet_videomem_2019, MemorabilityModeling2020, 2022MedievalEEGMem}. Alternatively, some studies use free recall for autobiographical memory triggered by media \cite{Bernd_mementos}.

In spontaneous conversation contexts, recognition tasks are impractical due to the variable content of free-flowing conversations. Instead, memory encoding and retention studies find free-recall self-reports to be more suitable as they allow participants to report memories without additional constraints \cite{benoit_participants_1996, diachek2024LinguisticFeaturesPredictRecall}. For example, these have been used in a recent behavioural study predicting memory encoding using linguistic features \cite{diachek2024LinguisticFeaturesPredictRecall}. For conversational memory retrieval datasets, other measures have been employed: a task-related memories survey in \cite{WoNoWA2020} and a third-party observer annotation of memory retrieval moments in \cite{Shaw2007homebirth}).

\subsection{Sample representation}
\label{diversity}
Lastly, participant samples vary in size and diversity. The sample size depends on the task context and length, ranging from 12 to 3246 subjects (\emph{Table \ref{fig:related_datasets}}). Considering demographics and representativeness of datasets' samples, three datasets do not report participants' demographics  \cite{videoMemorability2018, 2022MedievalEEGMem, MemorabilityModeling2020}, one dataset had students and university staff as participants \cite{WoNoWA2020} and one had female-only participants \cite{Shaw2007homebirth}. Only two out of seven memory-related datasets had a more balanced sample (except for the bias towards US residents) \cite{Bernd_mementos, diachek2024LinguisticFeaturesPredictRecall}.

\section{Data collection procedure}
\label{procedure}

\begin{figure*}
    \centering
    \includegraphics[width=0.7\textwidth]{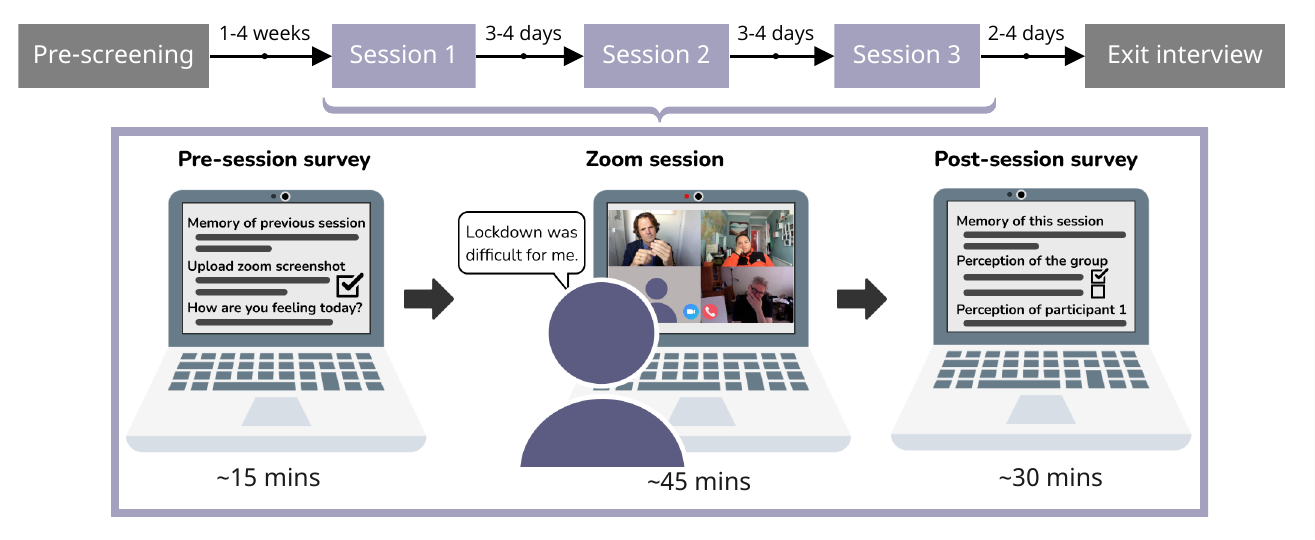}
    \caption{Experimental set-up. Upper flowchart - overall set-up. In the lower part - illustration of the procedure for every group session. On the second screen, there is a screenshot of a discussion from the MeMo corpus, except for the 4th participant shown with a person icon. The phrase the icon person produces is made-up but matches the conversations included in MeMo.} 
    \label{fig:setup}
\end{figure*}

We describe the \emph{MeMo} experimental procedure shown in \emph{Figure \ref{fig:setup}} in this section. We highlight how these relate to our stated primary goals (\emph{{G1}} and \emph{{G2}} described in \emph{Section \ref{motivation}}) and guiding principles (\emph{{P1}} to \emph{{P3}} described in \emph{Section \ref{principles}}). 

\subsection{Overall procedure} 
\label{overall_procedure}

The overall procedure of acquiring the \emph{MeMo} corpus is shown in \emph{Figure \ref{fig:setup}}. In this subsection we will guide the reader through the procedure step-by-step.

\subsubsection{Ethical approval.}
The Human Research Ethics Committee of TU Delft approved the \emph{MeMo} corpus data collection. Before the experiment, participants filled out an informed consent form permitting us to collect their personally identifiable data such as audio and video recordings to be later accessible to the research community under CC-by-NC license. Before the recordings participants were asked to come up with a pseudonym for themselves for the entirety of the recording. Participants were allowed to avoid answering questions if they did not want to and invent information about themselves throughout the recording.

\subsubsection{Conversation subjects.}
\textbf{Participants.} Participants were recruited using Prolific Academic recruiting service \cite{prolific}. All participants were required to reside in the UK, fluently speak English and be ready for a video-call study (i.e. possessing a laptop with a working camera and a headset with a microphone). The UK residency requirement served two purposes: first, the Prolific platform provided a large number of UK participants; second, research shows memory can be influenced by shared experiences \cite{mckinley_memory_2017}. With Covid-19 selected as a discussion topic, focusing on UK residents helped control for differences in pandemic experiences, thus enhancing the validity of memory comparisons between groups (\emph{$\rightarrow$ G1}).

To simulate a situation where facilitation is needed, such as in the case of differing opinions and perspectives in the group, we tried to maximise the diversity in opinions on our target topic - Covid-19 pandemic  (\emph{G2}). We, thus, targeted specific demographics differently affected by the pandemic. For example, parents of young children had to suddenly home-school their children or business owners faced work-related challenges having to adjust their business to changing regulations. The targeted groups were the following: parents with young children, older adults (50+), students, business owners. Five prescreening surveys were conducted on Prolific, each tailored to one of these groups (see \emph{Appendix 3}). Participants could only select one prescreening survey to avoid duplicate inclusion. Additionally, we ensured gender balance in our sample (\emph{$\rightarrow$ P1.3}).

\textbf{Group composition.}
The \emph{MeMo} corpus is designed around small-group discussions, typical of both work and informal settings, to ensure ecological validity (\emph{$\rightarrow${P1.1}}) and simulate potential facilitation scenarios (\emph{$\rightarrow${G2}}). "Small groups" here refers to 3 to 8 participants, an optimal size for allowing everyone to share their thoughts \cite{Wheelan2009GroupSize}. Participants completed a prescreening survey specifying their availability, and 5 to 8 participants were recruited per group, ensuring representation from each demographic. The minimum of 5 was set to maintain at least 3 participants per group, accounting for no-shows and dropouts. All participants were \textbf{zero acquaintance}, meaning they had never met before, allowing us to study the development of within-group relationships with no prior interactions \cite{Samp2007_FriendsMemory}. The group composition stayed the same throughout the experiment (except for having fewer members in later sessions if a participant dropped out after the first or the second sessions). Each participant took part in only one group, to avoid participant confusing memories from interactions with different groups. 

\textbf{Moderators.} For facilitation purposes (\emph{$\rightarrow${G2}}), discussions were guided by professional moderators who ensured a safe and inclusive environment, encouraging free-flowing conversation with spontaneous turn-taking (\emph{$\rightarrow${P1.2}}). Moderators were confederates aware of data collection goals and were allowed to use any methods of their liking for the facilitation of a free-flowing conversation. Moderators were not familiar with any participants before the experiment. Each group was assigned one moderator for the entirety of the experiment (3 sessions and an exit interview). Along with guiding the sessions, moderators had to fill out the same surveys as participants before and after each session, including memory reports and all other measures.

\subsubsection{Pre-screening survey.} The prescreening questionnaire included the consent form, participants' demographics (participants' age, gender, employment status, English fluency, country of residence), personality \cite{Hexaco_DEVRIES2013871}, values\cite{Schwartz_vals_2005}, technical requirements and online meeting experience (see \emph{Appendix 1}). Personality was included as it influences recall, with extroverts recalling more positive memories than those higher in neuroticism \cite{Mayo1989PersonalityMemory}. Values were also included, as they can enhance recall accuracy for items related to personal values \cite{Villaseñor2021ValuesMemory}.

\subsubsection{Pre-session survey.} Questionnaire data was collected using Qualtrics X platform \cite{qualtrics} (for the full content of the questionnaires see \emph{Appendix 1}). Participants started each session by completing a pre-session questionnaire that took $\sim$15 minutes to complete. The pre-session survey included the participant's mood assessment \cite{broekens2013affectbutton} before all sessions, as mood can affect memory encoding \cite{Jd1983MoodRecall, Matt1992Mood-congruent}. It also included long-term memory retention task (see \emph{Section \ref{long-term_task}}) before all the sessions except for the first one. Before the exit interview, there were some extra questions added to the pre-session survey - participants had to report a moment they found most important in all the past interactions and provide feedback on the moderator facilitation skills (see \emph{Section \ref{exitinterview}}for more detail). At the end of the survey, participants joined a scheduled Zoom link for the discussion session. They completed this survey by uploading a screenshot of their Zoom layout when all the participants were present in the Zoom session to facilitate eye-gaze target extraction (see \emph{Section \ref{other features}}). 

\subsubsection{Conversation session.}
All discussion sessions happened online, through a Zoom video-call platform, a typical software used for video-calls. This ensured that participants are in the comfort of their own homes, rather than in the lab (\emph{$\rightarrow${P1.1}}). They used their own computer and headsets, having their natural lighting, which also added to how comfortable they felt as well as the naturalistic setting of video-call discussions. To ensure that we can track participants' eye-gaze, the moderator asked participants to keep zoom in 'gallery' mode so that all the participants are on the screen at the same time and their location on the screen stays the same throughout each session. In addition, for a secure recording, there was a technical assistant involved in the call (with no camera or microphone on and no interaction with participants), who recorded the session and resolved any arising technical issues. 

After making sure that all participants had completed the pre-session survey, the moderator (or the technical assistant) would start the recording. The session began with head pose and gaze calibration for automatic post-experiment gaze direction annotation (see \emph{Section \ref{other features}}). Participants, guided by the moderator, first rotated their heads and then looked at each named participant on their screen. 

After that, guided by their moderator, participants started the discussion. To ensure a natural yet directed conversation (\emph{$\rightarrow${P1.2}}), the Covid-19 pandemic was chosen as a relatable topic, relevant to participants worldwide at the time of data collection (the year 2021), with diverse experiences and opinions. At the start of the first session, participants were informed that they would discuss the past, present, and future of the pandemic over three sessions, aiming to design a better future in case of a recurrence (the memory study focus was disclosed only at the experiment's end to prevent priming participants' memory). Each session lasted 45 minutes, providing sufficient time for conversation to emerge. This duration aligns with typical meeting and facilitation session lengths (\emph{$\rightarrow${G2}}, \emph{$\rightarrow${P1.1}}) \cite{Hayne1999GroupSupportFacilitation}. 

To reflect real-world conversations that repeat over time (e.g. work meetings), the corpus included 3 sessions spread out over 3-4 days, reflecting the frequency of real-world facilitation sessions, occurring once or twice a week \cite{Hayne1999GroupSupportFacilitation}. This longitudinal approach aimed to capture the evolution of participant relationships and conversational memory trends over time. (\emph{$\rightarrow${G1}},\emph{$\rightarrow${G2}}). This setup also provided repeated measures of memory at different points, capturing both short-term and long-term memory reports  (\emph{$\rightarrow${G2}}).

\subsubsection{Post-session survey.}
At the end of the Zoom session, the moderator reminded participants to open up the post-session survey link and start the questionnaire. After that, the recording stopped and the Zoom session was closed. See the summary of all measures used in the post-session survey \emph{Appendix 1}. The post-session survey started with free-recall self-reports (described in \emph{Section
\ref{free-recall_task}}) and a qualitative question for facilitation application (\emph{Section
\ref{facilitation_memory}}).
Since interpersonal skills can be of effect \cite{Miller_Interpersonal_2002}, participants' communication skills were evaluated by having participants rate each other's listening and conversational abilities with 2 one-item questions (see \emph{Appendix 2} for question formulations). Several scales were collected to measure relational growth and mutual understanding for the ultimate facilitation application of the \emph{MeMo} corpus (\emph{$\rightarrow${G2}}). Relational development was measured using the IOS scale \cite{aron1992IOS} to assess perceived closeness and a single-item scale was used to assess personal attitude (see \emph{Appendix 2}). Mutual understanding was assessed by comparing participants' pre-reported values with others' post-session evaluations using the Short Schwartz’s Value Survey \cite{Schwartz_vals_2005}. In addition, group perception was tracked with measures of cohesion \cite{braun2020cohesion}, entitativity \cite{koudenburg2014entitativity}, perceived interdependence \cite{gerpott2018interdependence}, situational characteristics \cite{DIAMONDSrauthmann2016ultra}, syncness, and rapport, as these factors can influence learning and group performance \cite{Evans1991CohesionPerformance, Kim2020CohesionLearning}.%Lea2004Cohesion, Piper1983CohesionAndBonding
The post-session survey finished with encoded event annotation (\emph{Section \ref{timing_task}}), and reasons for remembering (\emph{Section \ref{reasons_task}}). After submitting the post-questionnaire, the participants had to wait 3-4 days for the next scheduled session (or exit interview in case of the 3rd session), and then repeat the procedure (pre-session survey $\rightarrow$ conversation session $\rightarrow$ post-session survey).

\subsubsection{Exit interview. \label{exitinterview}} 3-4 days after the final conversation session, there was a $\sim$15-minute exit interview with each participant. Similar to the conversation sessions, there was a pre-session survey before this Zoom session. The only difference was that participants were asked what was the most important moment for them in all the previous discussion sessions. They then discussed that moment with the moderator one-on-one and answered a list of questions on the topic of the required capabilities of a social robot supporting public discussions, especially about what such a robot should remember.

\subsection{Memory measures \label{sec:memory_annotation}}

A common practice in user internal state modelling is to rely on third-party annotations of the investigated internal state. This can oversimplify these states and skew model accuracy, neglecting the first-party perspective and potentially introducing bias. This challenge applies to memory encoding studies, in which, traditionally, the free-recall reports are traced back to the encoded event by third-party annotators \cite{benoit_memory_1990, diachek2024LinguisticFeaturesPredictRecall}.
Here, we describe our method that mitigates these issues by leveraging participants' self-reports and the first-party task of aligning the reports with specific segments of recorded interactions. This method is aimed to preserve validity and minimise bias, particularly crucial for datasets like \emph{MeMo} corpus aiming to accurately represent the memory content as well as the event that that memory might be based on within the recorded data (\emph{$\rightarrow${P2}}).

\subsubsection{Free-recall self-reports.\label{free-recall_task}} To minimise the bias towards external stimuli and type of memory events, we have used a traditional free-recall task for memory self-reports \cite{Cleary2018MemoryMeasures}. The free-recall task was the first task in the post-session questionnaire, completed immediately after the end of the session, to avoid interventions of any additional bias that could modify the memory. The task formulation was open-ended to account for any conversational events recalled (spoken information, participants' feelings, context details etc., see the exact question formulation in \emph{Appendix 2}). 
Participants were meant to report a memorised 'moment' in each field in their own words without a word limit. They could report from 3 to 10 moments (the maximum was set to avoid fatigue and leave time for answering the next survey questions). Participants could move to the next survey questions only if they did not remember more moments or if they had already reported 10 moments. This way, we tried to capture all the retained and currently accessible \cite{Tulving1966AvailabileAccessibleMem} memories (unless there were more than 10 moments to report). The idea of having participants report memory 'moments' aimed to capture the content of the memories as well as the way participants segment the continuous stream of perceived and self-produced social signals into specific memorable events \cite{Kurby2008MemorySegmentation}.

\subsubsection{Encoded event annotation.\label{timing_task}} Unlike in previous research \cite{benoit_memory_1990, diachek2024LinguisticFeaturesPredictRecall}, the participants themselves did the assignment of their self-reports to specific events that happened within the conversation. This way, we wanted to ensure that the self-reported memories were correctly assigned to the encoded event they referred to, ensuring the validity and the accuracy of the resulting memory measure (\emph{$\rightarrow${P2}}). 
Within this encoded event annotation task, participants were given a link to the interaction recording and had to write down the start and end time (minute and second) of each moment they reported in the survey before (the memory self-reports were quoted back to them). Participants had the freedom of scrolling through the video and did not have to re-watch the entire recording to avoid additional fatigue. They also had an option of leaving the timing blank in case they could not find it, the moment was related to an overall feeling of discussion or other kind of memorable moments that cannot be connected to a particular interval in the interaction. We ensured that the free recall reports cannot be modified at this stage (\emph{$\rightarrow${P2}}). The encoded event annotation task was presented to participants at the very end of the post-session questionnaire so that the other survey questions would not be affected by seeing the video of the interaction either.

\subsubsection{Reasons for remembering.\label{reasons_task}} In addition to free-recall reports, we asked participants about the perceived reason or motivation behind their memory of each reported moment (see question formulation in \emph{Appendix 2}). The perceived reasons for memory were aimed to capture information that might not have been described in the free-recall reports - about the underlying personal significance of the specific moment and the underlying thought process as opposed to details of the event itself from free-recall reports (e.g. if the memory report is "I remember participant 3 said that they suffered from the lockdown", the reason could be "I remembered this moment because I also found it very difficult"). This information helps uncover intrinsic motivations for memory, useful for qualitative analysis and understanding in meeting facilitation (\emph{$\rightarrow${G2}}). Participants could provide as much detail as they wanted, and the question was placed at the end of the questionnaire to avoid biasing their recall.

\subsubsection{Long-term memory retention.\label{long-term_task}} Apart from the free-recall task immediately after the interaction (see above), the \emph{MeMo} corpus also included the measure of long-term retention. This measure was collected to investigate what kind of memories stay after the interaction and which memories are more likely to be forgotten (or less accessible for retrieval) (\emph{$\rightarrow${G1}}). To assess long-term memory retention, participants returned 3-4 days after the interaction, just before the next session, to answer free-recall questions about the previous session. This interval was chosen because most forgetting occurs within this timeframe, after which memory stabilizes, as shown by Ebbinghaus (1880) and later confirmed by Murre et al. (2015) \cite{Murre2015ForgettingCurve}. Therefore, the task meant to capture a stable representation of what participants' would remember in the long-term. The long-term memory question was exactly the same as the post-session free-recall task (see above). Similar to the original free-recall task, participants could report from 3 to 10 moments in text description fields with no word limit. Unlike the short-term annotations, this time the participants did not have to re-watch the video and map the timing to each moment, to avoid excessive fatigue before the conversation session.

\subsubsection{Qualitative data for facilitation application. \label{facilitation_memory}} Since the models trained on the \emph{MeMo} corpus are aimed to be applied to automatic meeting facilitation (\emph{$\rightarrow${G2}}), there was one qualitative question about what participants would want such a system to recall in the next sessions (for the task formulation see \emph{Appendix 2}.

\section{Dataset processing and curation}
In the following we describe distinct processing and filtering steps that we have applied to the raw dataset, resulting eventually in a curated version for the purpose of analysis and eventual sharing with the research community (see \emph{Section \ref{data_release}} for more detail).

\begin{figure*}
    \centering
    \includegraphics[width=\textwidth]{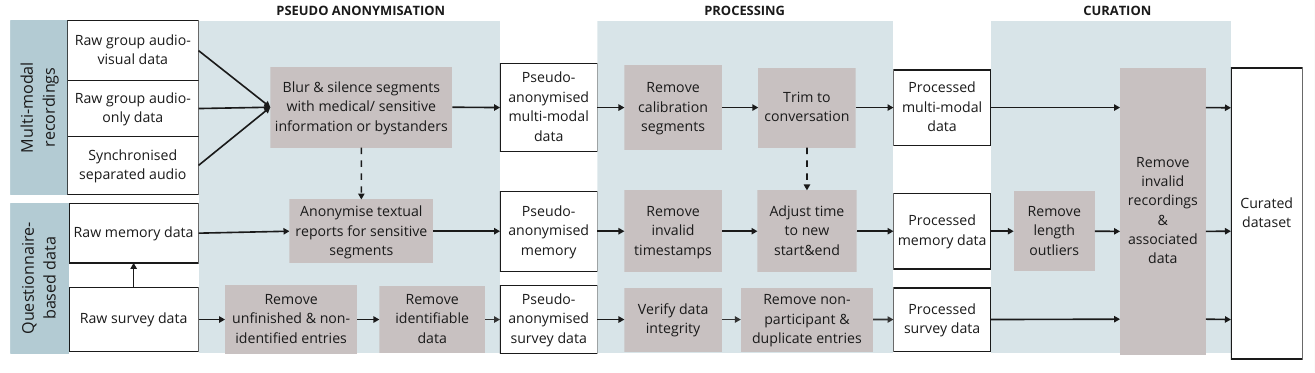}
    \caption{MeMo corpus processing and curation steps}
    \label{fig:curation}
\end{figure*}

\subsection{Pseudo-anonymisation}
\label{anonymisation}
To maintain the privacy of participants and compliance with ethical guidelines the dataset is being reviewed and processed to remove potentially problematic segments resulting in a pseudo-anonymised intermediate version, see \textit{Figure \ref{fig:curation}}.

\textbf{Multi-modal data pseudo-anonymisation.}
The raw multi-modal data recorded throughout the data collection consisted of group audio-visual recording, a group audio-only file in better quality, and separated audio channels per participant, automatically recorded through the Zoom software \cite{Zoom}. Because of a bug in the Zoom software, the separated audios were not aligned with the video recording. They, therefore, were first synchronised to align with the group audio-visual and audio-only recordings using the procedure described in \emph{Appendix 4}. 

Two types of data needed to be removed from the multi-modal recordings for privacy concerns. First, since the corpus contained discussions on the topic of Covid-19, the participants sometimes mentioned medical information about themselves or people close to them. Second, because of the recording being in a natural environment of participants' homes, sometimes there were bystanders passing by or talking in the background. Since they did not consent to be recorded, visible bystander faces and audible decipherable bystander speech needed to be removed. While we removed these types of data, the pseudo-anonymisation process is yet to go another stage of sensitive data removal before it will be shared (see the data release statement in \emph{Section \ref{data_release}}).

\textbf{Questionnaire data pseudo-anonymisation.} In regards to questionnaire data, the unfinished and non-identified entries were removed to only contain entries from identified paid participants. Identifiable data associated with the data collected through the Qualtrics survey platform was then removed. This included such data as IP address, location and signatures.  The Prolific IDs were replaced with a non-identifiable hash number, since they otherwise could be tracked to a specific account on the Prolific recruitment platform. After extracting participants' on-screen location via screenshots for further eye-tracking, the screenshots were also removed from the data since they were taken before the recording and, therefore, sometimes contained sensitive information such as participants' real names.

\textbf{Memory data pseudo-anonymisation.} Memory data extracted from the questionnaire data was pseudo-anonymised in accordance with the segments removed in multi-modal recording pseudo-anonymisation. For these sensitive segments, the textual reports of the corresponding memorable events were manually edited to avoid direct references and descriptions of the sensitive information. No free recall reports were removed, but the references to the sensitive information were replaced with "[anonymised]".

\subsection{Data processing}
As shown in \emph{Figure \ref{fig:curation}}, we have performed additional processing steps to improve the usability of the data.

\textbf{Multi-modal data processing.}
We processed the multi-modal data to keep only the conversation content, removing any technical or organizational parts from the recordings. Specifically, calibration segments for eye-gaze and head-pose, included at the beginning of each recording, were removed. We also trimmed the start and end of each recording to ensure they captured only the actual conversation time, excluding moments where moderators discussed technical or scheduling issues with participants or waited for participants to complete questionnaires. 

\textbf{Questionnaire data processing.}
The questionnaire data underwent the verification of integrity process. This constituted manually verifying the correspondence of session, group and alias information to the questionnaire data, since sometimes participants made mistakes in those fields. These were manually verified using the available information such as the date and time of questionnaire completion, the missing participants and the fields that were filled in correctly. In this processing step, only data for conversation participants was maintained, excluding participants who did not show up to any conversation sessions. The duplicate entries were also removed at this stage. Since sometimes participants forgot to fill in the surveys, there are some gaps in the data. To identify the consistency of the available data across groups, we computed a ratio of participants that completed all surveys (both pre- and post- in all sessions), we called that measure "questionnaire completeness" (see full table for all the groups in \emph{Appendix 5}). Overall, 13 out of 15 groups had questionnaire completeness at 50\% and above. We leave the decision of whether to discard some of the groups using this measure up to the future users of the corpus, since this might not be important for researchers not focusing on the longitudinal component of the corpus.

\textbf{Memory data processing.}  We excluded the memory moments with invalid timestamps: if the reported start was further than the reported end of a memory event and if the reported timestamps were outside the duration of the recording. Since the multi-modal data processing included shifting the start and the end of the recordings, the encoded event annotations needed to be adjusted to the new start and end of the recordings, to maintain the references to the originally tagged events.

\subsection{Data curation}
In addition to the necessary data processing, we have curated the processed data to provide the cleanest version of the dataset, which we recommend for further use (see \emph{Section \ref{data_release}} on how this data will be released). 

\textbf{Audiovisual data curation.}
Within the curation, we have removed a video that was recorded in 'speaker view', with one active speaker in the camera view at a time. All other videos were recorded in 'gallery mode', meaning that all participants were visible at all times. Given this criteria, 1 full session was removed.

\textbf{Questionnaire data curation.}
The questionnaire data remained unchanged except for removing the data associated with the removed session within audiovisual curation.

\textbf{Memory data curation.} 
Apart from removing memory data associated with the removed videos, within the curation, we have removed some memory event outliers. Specifically, we have removed memory moments outliers by duration: remembered events that lasted more than 1 standard deviation away from the mean duration over all the data (longer than 690 seconds). These moments were removed since they did not have enough detail or contained an overall feeling over the discussion rather than a specific event, in case of duration outliers (e.g. a moment with memory report "The agreement on the uncertainty of following rules and what rules were correct to follow etc." lasting 34 minutes). We also considered removing events shorter than 5 words in the free-recall memory description, but removing the length outliers also removed reports shorter than 5 words, since the duration of the associated event was always above the outlier threshold.

\subsection{Extracting multimodal features \label{other features}}

In addition to dataset processing and curation, we have extracted various features from the multi-modal data that can be useful for machine learning tasks.

\textbf{Transcription.} 
We diarised and transcribed the recorded audio of the discussions using the Kaldi Speech Recognition Toolkit \cite{Povey_ASRU2011}. We subsequently conducted manual reviews and corrections to the resulting transcripts where necessary. The timestamps for these transcripts are available at both the utterance and word levels, and we provide word-level transcriptions for each recording.

\textbf{Eye gaze.}
In our first corpus-based study, we used GazeSense software \cite{eyeware_2022} to estimate participants' gaze direction throughout the session. We created a customized grid for each participant, matching their Zoom gallery layout based on a screenshot they provided at the session's start. Each session began with a calibration where participants focused on specific screen segments. We used the coordinates of these segments as calibration points and then obtained gaze estimates for all frames beyond the final calibration. Due to recording imperfections and challenges with some participants' uploaded screenshots, gaze tracking for 13 participants was not possible. Therefore, our dataset includes eye gaze data from 40 participants from 14 groups, which collectively spanned approximately 23 hours of video recordings. 
This annotation includes gaze targets derived from screenshots capturing participants' screen views.

\textbf{Prosody.} 
We extracted the eGeMAPS feature set from the default eGeMAPS configuration in the OpenSmile software for prosody analysis \cite{amos_openface_2016}.

\textbf{Body Pose.}
Body and hand keypoints were extracted using MediaPipe \cite{Lugaresi2019MediaPipeAF}.
Keypoint prediction was evaluated on cropped segments of the original video to
ensure only a single person was visible to eliminate the need for keypoint tracking.
The largest available model was used and the confidence threshold for retaining
the predicted keypoints was set to $0.5$.

\textbf{Facial Action Units.}
Facial action units and face keypoints for participants were estimated using the OpenFace Software \cite{baltruvsaitis2016openface}. As with the body pose estimation, these features were extracted 
using the cropped segments to ensure only a single face was present in the video.

\section{Dataset contents}
\label{data_content}

\subsection{Dataset subjects}
\textbf{Conversation participants.}
There were 53 participants in the experiment. The demographics of the resulting sample are shown in \emph{Table \ref{tab:demographics}}. The sample was balanced across genders (28 F, 25 M), included participants of various age groups (from 18 to 76 y.o), and employment statuses (see \emph{Table \ref{tab:demographics}}). The participants all spoke fluent English and were UK residents. Maximising the diversity of opinions, we have recruited demographic groups that were differently affected by the pandemic (see specific criteria and the resulting selection in \emph{Appendix 3}).
\begin{table} [!t]
\caption{Participants' demographics}
\label{tab:demographics}
\centering
\begin{tabular}{llcc}
    \hline
    \hline
    
                                  &    \textbf{Mean +- SD}         & \textbf{Min} &  \textbf{Max} \\
    \textit{\textbf{Age (y.o)}}     & 38.8  +- 15.1  & 18 & 76   \\    
    \hline
    \hline\\
                            & & \textbf{Male} & \textbf{Female} \\
    \hline
    & & \\
    \textbf{\textit{Demographics}}  & \textbf{Full-time employed}                        & 11   & 11     \\
    %\hline
    \textbf{} & \textbf{Unemployed}                      & 10   & 3      \\
    %\hline
    & \textbf{Part-time employed}              & 7    & 5      \\
    %\hline
    & \textbf{Business owners}                 & 2    & 8      \\
    %\hline
    & \textbf{Students}                        & 6    & 7      \\
    %\hline
    & \textbf{Parents of young children}       & 3    & 6      \\
    %\hline
    & & \\
    \textbf{\textit{First Language}} & \textbf{English} & 23   & 23     \\
     & \textbf{Other} & 2   & 5     \\
    \hline
    \textbf{Total}            &               & 25   & 28     \\
    \hline \hline

\end{tabular}
\end{table}

\textbf{Conversation groups.}
The curated dataset sample contained 15 groups. The groups contained 4 participants on average, with a minimum of 2 (in two sessions where the third participant dropped out) and a maximum of 5 (see \emph{Table \ref{tab:descriptive_stats}}). Overall, 49 participants took part in all three sessions, and 4 participated in two sessions skipping one session for unforeseen circumstances. 

\textbf{Conversation moderators}
Four moderators (3 M, 1 F; 24-45 y.o.) were recruited to facilitate the conversation sessions. All moderators had 2 or more years of professional facilitation experience. Three moderators facilitated 3 groups each, and one facilitated 6 groups.

\textbf{Technical support.}
It is important to note that apart from group participants and the moderator, within each Zoom session there was a technical assistant who kept the camera and the microphone off throughout the entire experiment, they were recording the session and the moderator could communicate to them in case of any technical issues in a private chat. There was no communication between the technical assistant and the participants.

\subsection{Multi-modal Recordings}
\label{audiovisual data}
The curated dataset contains 31 hours (111674 seconds) of group audio-visual recordings. \emph{Table \ref{tab:descriptive_stats}} summarises the curated distribution of conversation sessions (the calibration timing is excluded from all the shown durations). Since there were 15 groups with 3 sessions each, a total of 45 sessions were recorded. The average duration of a conversational session was 42 minutes (2538 sec). This said the duration of the conversation differed across sessions, with the first session being the shortest (35 minutes or on average) and the third being the longest (46 minutes on average, see \emph{Table \ref{tab:descriptive_stats}} for detail). This difference is connected to the fact that, by the 3rd session, participants needed less explanation and had fewer technical issues than in the first and the second sessions.

\begin{table}[!t]
    \centering
    \caption{Descriptive statistics of participants' age, group size, conversation duration and memory reports}
    \label{tab:descriptive_stats}
    \begin{tabular}{l l c c c c}
    \hline
\hline
                              &                                             & \textbf{M}     & \textbf{SD}    & \textbf{Min}   & \textbf{Max}   \\
    \cline{3-6} 
    
    \textit{\textbf{Group size} }& & 3.7   & 0.8   & 2     & 5   \\ \cline{3-6} \\
    \textit{\textbf{Conversation}} & \textbf{Session 1}                                               & 2156 & 377  & 1290 & 2775 \\
    %\hline
    \textit{\textbf{duration}} & \textbf{Session 2}                                                & 2671 & 314  & 2160 & 3315 \\
    %\hline
   \textit{\textbf{(sec)}}   & \textbf{Session 3}                                                & 2761 & 349  & 1950 & 3260 \\
    %\hline
                              & \textbf{All sessions}                                                   & 2538 & 431  & 1290 & 3315\\ \\
    \cline{3-6} 
    \textit{\textbf{Memory}}                  & \textbf{Moment}                     &            &   &   &       \\ & \textbf{duration (sec)}                           & 141   & 183   & 1     & 1260  \\
          & \textbf{Moment count}                &             &    &    &          \\
%\hline
          & \textbf{per person}   & 3.9   & 1.4   & 1     & 10 \\ 
%\hline
              & \textbf{Word count}       &&&& \\ 
           & \textbf{per moment}     & 32    & 21    & 5     & 117 \\   
    \hline \hline 
\end{tabular}
\end{table}

\textbf{Video.} The videos were recorded with Zoom local recording. The video files have a sample rate of 32000 Hz with 32 bits per sample. The resolution is 1280x720, with a frame rate of 25. The duration of the video recording corresponds to the overall duration of the conversation session:  31. There are 45 videos in total - a video per session. 43 recordings were recorded in the 'gallery view' of Zoom with all participants on the screen, 2 recordings had some technical issues and were recorded in the 'speaker view' and therefore might need to be excluded from video analysis.

\textbf{Audio: Full and separated.}
Each video is accompanied by a separated full audio in .m4a format. This audio includes full audio from the above video automatically recorded through Zoom. This audio is available for all 45 sessions.
In addition to the full audio, the dataset includes automatically recorded audio channels per participant available for 42 out of 45 recorded sessions.

\textbf{Qualitative variations in multi-modal recordings.}  Since the dataset was recorded in the natural environment of Zoom conversations, the video data has some qualitative variation. This applies to the position the participants were in throughout the recording: while most participants were seated behind their desks, some participants were seated with a laptop on their lap and, in rare cases, a participant was lying down throughout the recording.  Another variation was the participants' location of taking the video call - while most participants were in the comfort of a home, one participant was seated outside, and two were taking a call from their car. In addition, although we asked the participants to keep their background as it is, some participants had a blurred background setting and, in rare cases, they had a virtual background. Another variation was the device used for the call - although a laptop was required, some participants joined from their tablet or a phone because of technical issues with their laptops (to our knowledge, this happened in 3 sessions). The final variation to possibly consider is the fact that participants used their own technical setup, with different quality microphones, cameras and internet connection. For the same reason, the lighting conditions might vary across participants.

\subsection{Questionnaire-based data}
%From the number of collected measures described in \emph{Section \ref{questionnaires}}, in this section, we describe the most relevant ones. 

\subsubsection{Longitudinal completeness of questionnaire data}
Although participants were asked to complete a questionnaire before and after each conversation session, there are some gaps in the data where participants could not complete questionnaires due to unforeseen circumstances. To facilitate the selection of the data for longitudinal analysis, we have computed the continuity of the questionnaire data in each group. Apart from the two groups, every group had 50$\%$ or more participants with complete pre- and post-questionnaires in all the sessions. We share the questionnaire completeness scores for each session for an easier subset selection process in future research (see table in \emph{Appendix 5} for the full table and measure description).

\subsubsection{Memory data}
\label{memory data}
The descriptive statistics of the curated memory annotation are shown in \emph{Table \ref{tab:descriptive_stats}}. The curated memory data included 602 moments reported in participants' free-recall tasks. The free recall description varied in length with an average of 32 words per reported moment. The mean duration of the self-reported memorable moments was 141 seconds (2.35 min), with a minimum of 1 second and a maximum of 1260 seconds. Participants remembered $\sim$4 moments on average (SD=1.4), with 1 as a minimum (in cases where participants didn't remember more or reported non-temporally attached moments) and 10 as a maximum. 

\begin{figure*}
    \centering
    \includegraphics[width=0.9\textwidth]{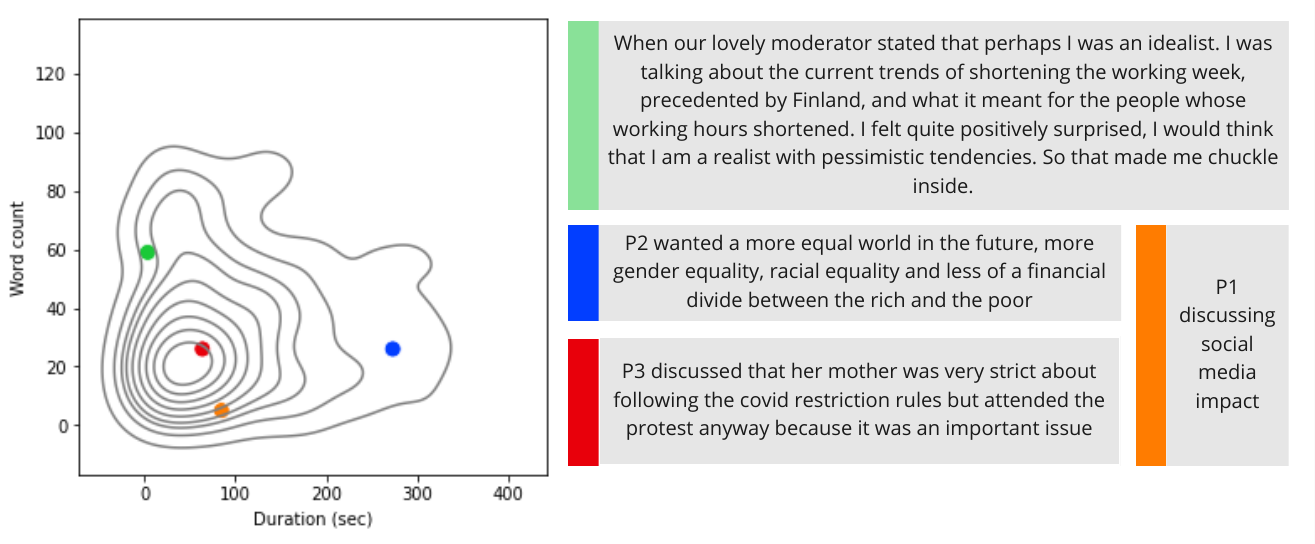}
    \caption{Duration and word-count distribution of memorable moments with examples of memorable moments shown with points and full reports on the right.}
    \label{fig:duration&wordcount}
\end{figure*}

There is a variability in memory description word count as well as the reported moment duration.  \emph{Figure \ref{fig:duration&wordcount}} shows examples of memorable moments variable in report length and moment duration. The memory description word count does not correlate with the duration of the moment annotated in the video (Spearman R=0.04 and p$>$0.05). This can be illustrated with the moment associated with the green point (top report) in \emph{Figure \ref{fig:duration&wordcount}}, with a long description of a memory associated with a moment that lasted for 4 seconds in the video recording.

\subsubsection{Other questionnaire measures}
Regarding other questionnaire measures, the collected self-assessment measures show high variability. 
For instance,  participants' mood was measured before each session with AffectButton \cite{broekens2013affectbutton}, assessing participants' pleasure, arousal and dominance from -1 to 1. The mood dimensions had a mean of 0.35  $\pm$  0.38 for Pleasure, -0.14 $\pm$ 0.77  for Arousal and 0.21 $\pm$ 0.57 for Dominance across all participants and sessions. In other words, participants' moods were generally minor positive, less aroused than average and slightly more dominant than average. The low arousal might be connected to the general setting of at-home video calls, illustrating the informal in-the-wild setting of the dataset (in accordance to P1.1, \emph{Section \ref{principles}}). This said, the variation in participants' mood was quite high as shown by high standard deviations across all scales. This might indicate, that the dataset might be usable for analysis of conversational memory in connection to the mood at the start of the conversation. 

Other self-reported measures, such as situation, group and relationship perception also show high variability. See \emph{Section \ref{session_dependency}} for more detail on these measures.

\subsection{Data release statement \label{data_release}} 

The MEMO dataset will be made publicly available following a comprehensive review process. This process aims at the removal of sensitive and potentially harmful information to the best of our knowledge. Once this step is complete, the wider release will occur gradually through multiple subsets of data, each focusing on different aspects of the dataset. The first release, associated with this paper, will include audio-visual recordings and temporal memory segment annotations. Unfortunately, specific timelines for the dataset release cannot be provided at this time. For any information on access to the dataset please contact Catharine Oertel.

\section{Using MeMo corpus for computational modelling}
\label{validity}

\subsection{Dependency Analyses}
%In this section, we describe potential variability in the corpus that is important to consider when it comes to computational modelling.

\subsubsection{Temporal dependency}
Conversational data intrinsically contains temporal dependencies. These and time-related biases related to human cognition have to be accounted for in the design of the computational models.

In relation to memory labels, there might be evidence for a particular time-dependent bias. Specifically, humans tend to recall the first and the last events from a sequence, a phenomenon referred to as recency/ primacy bias \cite{Cohen1982Primacy}. This, however, has not been investigated in the context of long interactions (rather than word lists and media-watching recall). If this hypothesis applies to the long discussions, most reported moments would occur in the beginning or the end of the session, and less in the middle of the session.

To test the hypothesis, we compared the memorability index (percentage of participants that included the segment in their memory reports, see labels used in \cite{Tsfasman2022}) of moments that occurred in the first 1/3 of the session, in the middle and in the end 1/3 of the session. Judging by an ANOVA followed by a paired t-test, the memorability index of the moments at the start and the end of the conversations is significantly higher than the ones in the middle (p$<$0.005). This therefore seems to confirm the recency and primacy bias in relation to memorable moments occurrence. Therefore, the memory labels should not be treated as independent of temporal context within a session.

\subsubsection{Session dependency \label{session_dependency}}

Another factor to consider in the \emph{MeMo} dataset relates to its' longitudinal quality. Since each group participated in 3 consecutive sessions, starting as strangers and gradually getting to know each other, there might have been some evolution in their relationships and group perception. This matters for two reasons. First, a session-dependent variation would indicate the success of moderated sessions in creating a connection within the group (\emph{$\rightarrow$ G2}) and show whether the \emph{MeMo} corpus is representative of general societal trends. Second, this variability is important for computational modelling to determine whether the sessions can be treated as independent of each other. To investigate whether there is a consistent change in participants' perceptions of each other and the group, we analysed self-reported ratings collected from participants after each session during the experiment. 

On the \textbf{group level}, to investigate the development of interaction and group perception, we compared the questionnaire ratings task across sessions. In case the facilitation sessions were successful, the hypothesis is that there would be higher ratings of group cohesion \cite{braun2020cohesion}, entitativity \cite{koudenburg2014entitativity}, rapport and syncness in the consequent sessions in comparison to the first one. 
The significance of the differences was evaluated with Friedman chi-square as a non-parametric analogy of repeated measures ANOVA (since the assumption of normality was not met by the data). The comparison of group perception scores between different sessions showed that the hypothesis is confirmed in relation to syncness (Friedman $\chi^2$=13.7, p$<$0.005), rapport (Friedman $\chi^2$=10.1, p$<$0.005) and entitativity (Friedman $\chi^2$=31.7, p$<$0.005). In other words, participants considered that the group was significantly more harmonious, in sync and united in the 3rd session in comparison to the 1st session. This, however, did not hold for the group cohesion measure, with insignificant differences between the 1st and the 3rd sessions (Friedman $\chi^2$=5.4, p$>$0.05).

\begin{figure}
    \centering
    \includegraphics[width=0.5\textwidth]{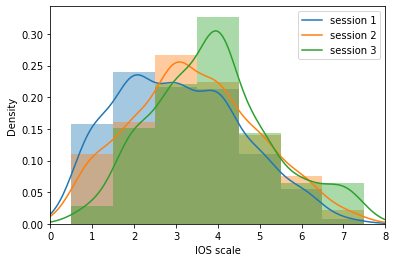}
    \caption{The change in perceived social distance between participants throughout the 3 sessions of the interactions, reported through IOS scale, with 1 = no overlap, 3 = some overlap and 7 = most overlap \cite{aron1992IOS}.}
    \label{fig:IOS}
\end{figure}

On the level of \textbf{individual relational development}, to show whether there was a similar development in how close participants felt to each other throughout the three sessions, we investigated the change in IOS scores \cite{aron1992IOS}. IOS (Inclusion of Other in Self) scale is a validated and comprehensible measure used to evaluate perceived relationship closeness between two participants \cite{Gächter2015IOSClosenes}. Using this scale, each participant in \emph{MeMo} evaluated how close they felt to every participant in their group after every session. The scale implies that the more familiar the participants felt with each other, the higher they would rate their subjective proximity on IOS scale. The hypothesis was that, with every next group session, participants would feel closer to each other, indicating growing closeness between participants. After comparing participants' IOS scores between sessions, the Friedman chi-square test showed that, indeed, participants felt closer to each other with every subsequent session (Friedman chi-square = 331.3, p$<$0.005). \emph{Figure \ref{fig:IOS}} illustrates how the distribution of IOS assessment gradually moves from the mode of 2 (little overlap) in the 1st session to 4 (equal overlap) in the 3rd session. This is consistent with previous findings showing that less acquainted participants score an average of 2 on the IOS scale, with friends scoring about 4 \cite{Gächter2015IOSClosenes}, therefore showing that the majority of participants developed a friendly relationship after the 3 discussion sessions facilitated by professional moderators in the \emph{MeMo} dataset.

These results confirm that the \emph{MeMo} corpus setup oversees the development of interpersonal relationships throughout the 3 interactions and therefore can serve as a useful tool for the evaluation of research questions related to longitudinal change in group dynamics and interpersonal relationships throughout repeated deliberation sessions guided by a professional moderator. This also shows that the sessions cannot be treated as independent from each other, with systematic relational and group perception differences between consecutive sessions.

\subsection{Example: Group-level Conversational Memory Prediction \label{ML}}

The first baseline for memorability prediction based on the \emph{MeMo} corpus has been explored and shown to be promising in \cite{Tsfasman2022}. Tsfasman et al. \cite{Tsfasman2022} used group gaze features to automatically predict group memorability levels. 

\textbf{Group memorability levels} were computed from the first-party memory annotations described in \emph{Section \ref{sec:memory_annotation}}, which provided the opportunity to calculate the percentage of participants that remembered each 5-second time-window of recorded discussions. Tsfasman et al. \cite{Tsfasman2022} used these proportions to create four labels of memorability level: 'zero' if no one from the group included a time window in their reports, 'low' if less than 30\% remembered that time slice, 'middle' - 30 to 70\% of participants considered a moment memorable, 'high' - if more than 70\% reported an interval in their annotations. 

\textbf{Eye gaze features}. To analyse group eye gaze behaviour and train the classifier, Tsfasman et al. \cite{Tsfasman2022} engineered multiple features computed from raw eye gaze direction and speaker activity annotation (see \emph{Section \ref{sec:memory_annotation}}).  Specifically, such group features as eye gaze presence, MaxGaze and Entropy. They also introduced two additional measures based on speech - the proportion of participants looking at the speaker at each given time slice and the proportion of participants speaking in the given time slice.

\textbf{Modelling.} Tsfasman et al. (2022) \cite{Tsfasman2022} trained a Random Forest classifier and a Multi-layer Perceptron on the \emph{MeMo} data with input being eye gaze features and output being four group memorability levels (zero, low, middle and high group memorability). Both models have been shown to predict the memorability labels above chance (chance being 0.25) with balanced accuracy scores of 0.42 and 0.43 respectively. These first results show how the \emph{MeMo} corpus can be used for both computational (prediction of group memorability levels) and qualitative (analysis of different categories of memorable moments) studies. Training a rudimentary classifier on non-verbal features to predict the memorability level of a conversational interval resulted in above-chance performance. For future work, there is a need to create more complex models that take into account the temporal and session dependencies of the data points.

\subsection{Potential future tasks \label{uses}}
As mentioned before, memory can be broken down into three subprocesses - encoding, retention and retrieval. Since memory representations are not directly observable, modelling all three processes relies on memory retrieval tasks (e.g. free-recall reports in our case) and the three subprocesses are never completely separable from each other. However, the main focus can be on one of the three. As such, we divide the possible memory modelling tasks by the primarily modelled memory subprocess (see \emph{Section \ref{fig:related_datasets}}).

\subsubsection{Conversational memory encoding modelling}
\textbf{Task Description.}
Modelling conversational memory encoding is the primary task of \emph{MeMo} corpus. Investigating the moment encoding, or preserving an event in memory, means focusing on the features of the timeframe of the actual event that a person's memory refers to.  The task involves predicting the likelihood of a conversational segment being encoded by the conversation participants. In other words, a computational model would be trained on (non-)verbal behaviour of the participants at the moment of either encoding or not encoding a specific segment.
 
A predictive model could infer the likelihood of encoding an event either by individual participants (human-centred approach) or by the group as a whole (situation-centred approach). In a human-centred approach, the model could focus on individual behaviours such as eye gaze, gestures, and speech—in relation to their own memory rating, investigating what behaviours indicate that a moment is important enough for a person to encode. Other participants' behaviours can also be used to understand what conversational context makes a moment memorable for an individual. In contrast, a situation-centred approach could focus on what qualities make a moment "universally" memorable for the group as a whole. Here, the model could use the behaviours of all participants, with the output being a cumulative metric of the percentage of participants who successfully encoded that specific moment (see \emph{Section \ref{ML}} or \cite{Tsfasman2022} for an example).

\textbf{Task Relevance.} 
Modelling how conversational moments are encoded through participants' (non-)verbal behaviour could be useful for various applications. In meeting facilitation systems, identifying which segments are likely encoded by participants can help focus on moments that enhance mutual understanding and shared history \cite{Garon2002FacilitatingMeetings, Clark2019CommonGroundNeedCA}. Sharing this information might highlight points one person found important but others missed, prompting further discussion. In automatic meeting summarisation, recognising which segments are likely encoded can produce summaries that emphasise key information or enhance memory by focusing on less memorable points \cite{Wrede2003SpottingS, Nihei2019SummarisationImportance}. %This could be done in real time, offering new conversational input.

\textbf{MeMo Resources for the Task.}
The \textbf{output} labels for this task could be the memory features described in \emph{Section \ref{sec:memory_annotation}} - free recall reports combined with participants' encoded events annotation (see descriptive statistics of the memory measure in \emph{Section \ref{memory data}}). These memory data could then be divided into binary labels per chosen time slice (retained/ non-retained for each specific participant) or a cumulative group-level aggregated measure as described in \emph{Section \ref{ML}} or \cite{Tsfasman2022}.  One more possible option of an output variable could be a textual description from the participants' memory report(s). The raw video or audio (full as well as separated channels) data that can be used for the \textbf{input} measures is described in \emph{Section \ref{audiovisual data}}. Along with that, a variety of automatically extracted features is available for this purpose (see \emph{Section \ref{other features}}).

\subsubsection{Conversational memory retention modelling}
\textbf{Task Description.}
To model \textbf{how humans retain conversations on a long-term}, it is possible to use the \emph{MeMo} corpus with the long-term memory reports. Modelling retention implies predicting whether a moment will be remembered long-term using free-recall reports from two points in time: shorter term - straight after the interaction, and longer-term - 3-4 days after the interaction when most forgetting would have occurred and only the most persistent memories would have stayed \cite{Murre2015ForgettingCurve}. The task, in this case, is similar to the encoding modelling, except for more incremental output labels -  a time slice could be forgotten (not mentioned in any memory reports), retained in the short-term (only mentioned in the short-term memory reports), and retained long-term (mentioned in both short-term and long-term reports). For this purpose, however, more annotation is required to connect the long-term memory reports to the events in the recording. The input features could be the same as those in the encoding task - various verbal and non-verbal behaviour during the conversation.

\textbf{Task Relevance.} 
This task is relevant for long-term facilitation systems and conversational agents. Since long-term user engagement remains a challenge \cite{Akhter-Khan2020Why, Mitsuki}, efforts have been made to enhance agents with shared memory models \cite{elvir_remembering_2017, brom_episodic_2010, kasap_building_2012, Nash2022}. However, there is no way to identify which memories are shared or forgotten by the user. Memory retention modelling could improve agents' long-term understanding of users and adapt dialogue strategies based on the likelihood of events being retained. Facilitation systems could also benefit from knowing which events are retained long-term. These models could be personalised for better user experience or modified to support dementia patients.

\textbf{MeMo Resources for the Task.}
The output labels for this task could be the memory reports described in \emph{Section \ref{sec:memory_annotation}} - free recall reports, participants' encoded events annotation and long-term memory reports. The input features could be similar to the encoding task above (audiovisual data described in \emph{Section \ref{audiovisual data}} and features from \emph{Section \ref{other features}}).

\subsubsection{Perceived reason for retention modelling}
\textbf{Task Description.} To model the reasons why conversational segments were encoded and retained, the \emph{MeMo} corpus contains self-reported reasons why participants recalled each moment. The reasons were then categorised by two annotators with the label frequencies reported in \cite{Tsfasman2022}. In future research, it would be important to further investigate the types of reasons that participants report for considering a moment memorable. This could be a separate modelling task of inferring a perceived reason for remembering a moment from verbal or non-verbal data as well as memory reports themselves.

\textbf{Task Relevance.} 
This task offers a qualitative perspective on conversational memory modelling, providing an opportunity for deeper user understanding in applications such as meeting facilitation and conversational agents. Inferring and understanding the perceived reasons for remembering can help extract information about the underlying relevance of a specific memorable event for a user. This reasoning could be crucial for further dialogue strategies in human-agent interaction.
What is more, perceived reasons can serve to improve the accuracy of the encoding prediction, and to refine the encoding memory prediction labels into more specific memorable event categories, potentially reducing noise in the data.

\textbf{MeMo Resources for the Task.}
The output labels could be the description of the perceived reasons mentioned in section in \emph{Section \ref{sec:memory_annotation}}, as well as a categorisation of these descriptions described in \cite{Tsfasman2022} available by request. The input labels could be the free-recall memory reports as well as any conversational behaviour, similar to the encoding and retention tasks above.

\section{Conclusions}

In the present paper, we introduce \emph{MeMo} - the first multimodal corpus with first-party memory annotations. The multi-party interactions were conducted in an ecologically valid, spontaneous setting in a typical online meeting environment and participants in the comfort of their own homes. For each group, there were three 45-minute group meetings spread 3-4 days apart, which provided opportunities for investigating group dynamics and relationships in newly formed groups. The rich collection of perceptual measures from pre- and post-session questionnaires can serve as a source for studies on how individual participant characteristics and perception of interaction, group, and other participants affect group dynamics and conversational memory.
With the first investigation on how non-verbal signals can indicate conversational memorability, we show that group eye-gaze behaviour can discriminate over conversational memorability levels \cite{Tsfasman2022}. We also show that throughout 3 sessions there was an interpersonal relationship development: participants assessed their groups as having more rapport, syncness and entitativity. Participants assessed their social distance from other participants being closer in the 3rd session in comparison to the first session. 

With this research, we hope to pioneer multi-modal corpus research of conversational memory and create opportunities for studying conversational memory and group dynamics along with other topics related to longitudinal group discussions.

\section{Limitations}
\label{sec:limitations}
As with any dataset, the \emph{MeMo} corpus has its limitations.
In the following, we will highlight several key aspects relevant to the intended use of the corpus.

\paragraph*{\textbf{Validity of Memory Annotations}} \emph{MeMo} contains two types of memory annotations: Free-Recall Self-reports (FRS), and Encoded Event Annotations (EEAs). While we argue that jointly these provide valid resources for modelling both the encoding and retention processes of participants, there are some caveats we want to highlight.

First, our corpus assumes that each FRS corresponds to a single EEA, i.e., a remembered moment in the conversation corresponds to a single segment in the multimodal signal. However, this might not always be true, for example, when multiple similar events are reported as a single memory (with more than one EEA included in one FRS).

Secondly, it is likely that there are events that participants remembered, but failed to either report or annotate. Our protocol required participants to identify multiple segments in the recording at a fine granularity, which is a cognitively taxing task. This might impact the comprehensiveness and accuracy of the resulting annotations.

Finally, our protocol asked participants to report FRS from previous sessions (see \emph{Section \ref{long-term_task}}). We have argued that this data forms a viable source for modelling retention. However, we cannot rule out that participants could access memories from previous sessions but failed to report them (i.e., we do not know if they forgot about a moment or simply failed to produce a matching FRS).

\paragraph*{\textbf{Comprehensiveness of Memory Annotations}} While we believe that FRS and EEAs capture properties relevant for modelling retention and encoding, the current annotations do not facilitate work on the functional use of memories in conversations (i.e., how people use memories to convince others or for bonding purposes \cite{Bluck2005FunctionsMemory}). However, it is likely that \emph{MeMo} captures a substantial amount of such instances, and we intend to investigate their presence in future work (i.e., to provide additional annotations). Note that even though our protocol for obtaining FRS explicitly asks participants to "recall" memory, this is merely a methodological necessity (i.e., it is not possible to study memory content without prompting for it); it does not facilitate modelling retrieval processes as such (i.e., the conditions under which content is dynamically accessed or accessible).

\paragraph*{\textbf{Choice of Conversational Setting}}
First, the \emph{MeMo} corpus was recorded online. Therefore, it is specific to the memory of online video-call conversations, which could be different from the memory of face-to-face interaction, because of different conversational dynamics \cite{Tian2023TurntakingOnlineVSOffline}. This said, the research on how humans remember information from online and face-to-face lectures shows no differences in recall quality between these settings \cite{Ng2007OnlineRecallEqualsOffline}.

Second, the topic of the conversations in the \emph{MeMo} corpus is limited to the Covid-19 pandemic. At the time of the recording, this topic was naturally engaging since it noticeably affected most people's lives. Humans tend to remember information that is personally relevant to them \cite{Ray2019OMemoryImportance} and therefore with a different topic, the trends in memorable moments might have been different. 

Additionally, having a trained moderator may create a sense of hierarchy in the group as well as introduce different moderation styles, potentially affecting the discussion structure and group dynamics. A different setting might lead to different results, depending on the environment, goals, and roles in the conversation.

\section*{Acknowledgements}
We would like to thank Jose Vargas-Quiros for the fruitful discussions in October 2020 that led to the initial idea of the \emph{MeMo} corpus. We also thank Wendy Aartsen, Tiffany Matej, Hayley Hung, Pradeep Murukannaiah and Michiel van der Meer for all the advice on the design of the corpus.

This material is based upon work supported by Delft Institute for Values; Hybrid Intelligence Center, a 10-year program funded by the Dutch Ministry of Education, Culture, and Science
through the Netherlands Organisation for Scientific Research; European Commission funded project ``Humane AI: Toward AI Systems That Augment and Empower Humans by Understanding Us, our Society and the World Around Us'' (grant \# 820437); the National Science Foundation (NWO) under Grant No. (1136993) and Grant No. (024.004.022); TAILOR, a project funded by EU Horizon 2020 research and innovation programme under GA No 952215.
Any opinions, findings, conclusions or recommendations expressed in this material are those of the author(s) and do not necessarily reflect the views of the supporting institutions. The support is gratefully acknowledged.

\section*{Author Contributions}

The contributions in the work presented in this paper were the following (\href{https://credit.niso.org/}{CRediT taxonomy}). \textbf{Maria Tsfasman}: conceptualisation, methodology, resources, data curation, writing - original draft, writing - review \& editing, visualisation, project administration, software, formal analysis, investigation.  \textbf{Bernd Dudzik}: writing - original draft, writing - review \& editing, project administration, supervision. \textbf{Kristian Fenech}: conceptualisation, methodology, software, data curation, funding acquisition, writing - review \& editing. \textbf{Andras Lorincz}: conceptualisation, resources, funding acquisition. \textbf{Catholijn M. Jonker}: conceptualisation, methodology, writing - review \& editing, funding acquisition. \textbf{Catharine Oertel}: conceptualisation, methodology, resources, data curation, writing - original draft, writing - review \& editing, project administration, funding acquisition.

\bibliographystyle{IEEEtran} 
\bibliography{bib/vFinal_MeMo_TAC_2024.bib}  

\begin{thebibliography}{100}
\providecommand{\url}[1]{#1}
\csname url@samestyle\endcsname
\providecommand{\newblock}{\relax}
\providecommand{\bibinfo}[2]{#2}
\providecommand{\BIBentrySTDinterwordspacing}{\spaceskip=0pt\relax}
\providecommand{\BIBentryALTinterwordstretchfactor}{4}
\providecommand{\BIBentryALTinterwordspacing}{\spaceskip=\fontdimen2\font plus
\BIBentryALTinterwordstretchfactor\fontdimen3\font minus \fontdimen4\font\relax}
\providecommand{\BIBforeignlanguage}[2]{{%
\expandafter\ifx\csname l@#1\endcsname\relax
\typeout{** WARNING: IEEEtran.bst: No hyphenation pattern has been}%
\typeout{** loaded for the language `#1'. Using the pattern for}%
\typeout{** the default language instead.}%
\else
\language=\csname l@#1\endcsname
\fi
#2}}
\providecommand{\BIBdecl}{\relax}
\BIBdecl

\bibitem{Bluck2005FunctionsMemory}
S.~Bluck, N.~Alea, T.~Habermas, and D.~C. Rubin, ``A tale of three functions: the self–reported uses of autobiographical memory,'' \emph{Soc Cognition}, vol.~23, 2005.

\bibitem{NeuroEncyclopedia2021}
P.~Rotshtein, Ed., \emph{\BIBforeignlanguage{English}{Encyclopedia of Behavioral Neuroscience, 2nd edition}}.\hskip 1em plus 0.5em minus 0.4em\relax Netherlands: Elsevier, 2021.

\bibitem{mckinley_memory_2017}
G.~L. McKinley, S.~Brown-Schmidt, and A.~S. Benjamin, ``\BIBforeignlanguage{en}{Memory for conversation and the development of common ground},'' \emph{\BIBforeignlanguage{en}{Mem Cognit}}, vol.~45, no.~8, 2017.

\bibitem{benoit_participants_1996}
W.~L. Benoit, P.~J. Benoit, and J.~Wilkie, ``\BIBforeignlanguage{en}{Participants’ and observers’ memory for conversational behavior},'' \emph{\BIBforeignlanguage{en}{South Comm J}}, vol.~61, no.~2, 1996.

\bibitem{Samp2007_FriendsMemory}
J.~A. Samp and L.~R. Humphreys, ``"i said what?" partner familiarity, resistance, and the accuracy of conversational recall.'' \emph{Commun Monogr}, vol.~74, no.~4, 2007.

\bibitem{Knutsen2014EgoMemory}
D.~Knutsen and L.~Le~Bigot, ``Capturing egocentric biases in reference reuse during collaborative dialogue,'' \emph{Psychon Bull Rev}, vol.~21, no.~6, 2014.

\bibitem{Miller_Interpersonal_2002}
J.~B. Miller and P.~A. de~Winstanley, ``The role of interpersonal competence in memory for conversation,'' \emph{Pers Soc Psychol Bull}, vol.~28, no.~1, 2002.

\bibitem{diachek2024LinguisticFeaturesPredictRecall}
E.~Diachek and S.~Brown-Schmidt, ``\BIBforeignlanguage{en}{Linguistic features of spontaneous speech predict conversational recall},'' \emph{\BIBforeignlanguage{en}{Psychon Bull Rev}}, 2024.

\bibitem{poria2017AffectiveComputingReview}
S.~Poria, E.~Cambria, R.~Bajpai, and A.~Hussain, ``A review of affective computing: From unimodal analysis to multimodal fusion,'' \emph{Inf Fus}, vol.~37, 2017.

\bibitem{Vinciarelli2009SSP}
A.~Vinciarelli, M.~Pantic, and H.~Bourlard, ``Social signal processing: Survey of an emerging domain,'' \emph{Image Vis Comput}, vol.~27, no.~12, 2009.

\bibitem{cambria2014NLP}
E.~Cambria and B.~White, ``Jumping nlp curves: A review of natural language processing research,'' \emph{IEEE Comput Intell Mag}, vol.~9, no.~2, 2014.

\bibitem{Bender2021Parrots}
E.~M. Bender, T.~Gebru, A.~McMillan-Major, and S.~Shmitchell, ``On the dangers of stochastic parrots: Can language models be too big?'' in \emph{Proceedings of the ACM FAccT '21}.\hskip 1em plus 0.5em minus 0.4em\relax ACM, 2021.

\bibitem{nosek2022replicability}
B.~A. Nosek, T.~E. Hardwicke, H.~Moshontz, A.~Allard, K.~S. Corker, A.~Dreber, F.~Fidler, J.~Hilgard, M.~Kline~Struhl, M.~B. Nuijten \emph{et~al.}, ``Replicability, robustness, and reproducibility in psychological science,'' \emph{Ann Rev Psych}, vol.~73, 2022.

\bibitem{hutson2018AIreproducibility}
M.~Hutson, ``Artificial intelligence faces reproducibility crisis,'' 2018.

\bibitem{Paullada2021MLEcologicalVal}
A.~Paullada, I.~D. Raji, E.~M. Bender, E.~Denton, and A.~Hanna, ``Data and its (dis)contents: a survey of dataset development and use in machine learning research,'' \emph{Patterns}, vol.~2, 2021.

\bibitem{Holt‐Lunstad2021SocialConnectionsAndHealth}
J.~Holt‐Lunstad, ``The major health implications of social connection,'' \emph{Curr Dir Psychol Sci}, vol.~30, 2021.

\bibitem{Reis2000Relatedness}
H.~T. Reis, K.~M. Sheldon, S.~L. Gable, J.~A. Roscoe, and R.~M. Ryan, ``Daily well-being: the role of autonomy, competence, and relatedness,'' \emph{Pers Soc Psychol Bull}, vol.~26, 2000.

\bibitem{Garcia1991Mediation}
A.~Garcia, ``Dispute resolution without disputing: How the interactional organization of mediation hearings minimizes argument,'' \emph{Am Soc Rev}, vol.~56, 1991.

\bibitem{Picard2011DeepeningConversations}
C.~A. Picard and M.~Jull, ``Learning through deepening conversations: A key strategy of insight mediation,'' \emph{Confl Res Quart}, vol.~29, 2011.

\bibitem{Dillard2013DeliberationFacilitation}
K.~N. Dillard, ``Envisioning the role of facilitation in public deliberation,'' \emph{J Appl Commun Res}, vol.~41, 2013.

\bibitem{Garon2002FacilitatingMeetings}
J.~E. Garon, ``Facilitating meetings.'' \emph{Clin Leadersh Manag Rev}, vol. 16 4, 2002.

\bibitem{Bruce2017FamilyMeetingSupport}
C.~Bruce, A.~D. Newell, J.~H. Brewer, D.~O. Timme, E.~Cherry, J.~Moore, J.~Carrettin, E.~Landeck, R.~Axline, A.~Millette, R.~Taylor, A.~Downey, F.~Uddin, D.~Gotur, F.~Masud, and D.~Zhukovsky, ``Developing and testing a comprehensive tool to assess family meetings: Empirical distinctions between high‐ and low‐quality meetings,'' \emph{J Crit Care}, vol.~42, 2017.

\bibitem{Hayne1999GroupSupportFacilitation}
S.~C. Hayne, ``The facilitators perspective on meetings and implications for group support systems design,'' \emph{SIGMIS Database}, vol.~30, no. 3–4, 1999.

\bibitem{Phillips1993Faciliated}
L.~Phillips and M.~C. Phillips, ``Faciliated work groups: Theory and practice,'' \emph{J Oper Res Soc}, vol.~44, 1993.

\bibitem{Lindblom2009LargeMeetingSupport}
T.~Lindblom, M.~W. Aiken, and M.~Vanjani, ``Electronic facilitation of large meetings,'' \emph{Communications of the IIMA}, 2009.

\bibitem{Shamekhi2019RealtimeMeetingSupport}
A.~Shamekhi and T.~Bickmore, ``A multimodal robot-driven meeting facilitation system for group decision-making sessions,'' in \emph{ICMI '19}.\hskip 1em plus 0.5em minus 0.4em\relax New York, NY, USA: ACM, 2019.

\bibitem{Li2022VRTurntakingSupport}
J.~V. Li, M.~Kreminski, S.~M. Fernandes, A.~Osborne, J.~McVeigh-Schultz, and K.~Isbister, ``Conversation balance: A shared vr visualization to support turn-taking in meetings,'' in \emph{Extended Abstracts of the 2022 CHI EA '22}.\hskip 1em plus 0.5em minus 0.4em\relax New York, NY, USA: ACM, 2022.

\bibitem{Schiavo2014OvertGroupSupport}
G.~Schiavo, A.~Cappelletti, E.~Mencarini, O.~Stock, and M.~Zancanaro, ``Overt or subtle? supporting group conversations with automatically targeted directives,'' \emph{Proceedings of the 19th ACM IUI}, 2014.

\bibitem{Kulyk2005MeetingSupportAttention}
O.~A. Kulyk, J.~Wang, and J.~M.~B. Terken, ``Real-time feedback on nonverbal behaviour to enhance social dynamics in small group meetings,'' in \emph{MLMI}, 2005.

\bibitem{Kim2008MeetingSupportDominance}
T.~J. Kim, A.~Chang, L.~Holland, and A.~S. Pentland, ``Meeting mediator: enhancing group collaboration using sociometric feedback,'' \emph{Proceedings of the 2008 ACM CSCW}, 2008.

\bibitem{Nowak2023MeetingSupportSocialPr}
K.~Nowak, L.~Tankelevitch, J.~Tang, and S.~Rintel, ``Hear we are: Spatial audio benefits perceptions of turn-taking and social presence in video meetings,'' \emph{Proceedings of the 2nd CHI Work}, 2023.

\bibitem{Okada2019Affective}
T.~Okada, S.~Okamoto, and Y.~Yamada, ``Affective dynamics: Causality modeling of temporally evolving perceptual and affective responses,'' \emph{IEEE TAC}, vol.~13, 2019.

\bibitem{Norman2009MemoryVSExperience}
D.~A. Norman, ``The way i see it memory is more important than actuality,'' \emph{Interactions}, vol.~16, no.~2, 2009.

\bibitem{wixted_memory_2018}
N.~M. Long, B.~A. Kuhl, and M.~M. Chun, ``\BIBforeignlanguage{en}{Memory and {Attention}},'' in \emph{\BIBforeignlanguage{en}{Stevens' {Handbook} of {Experimental} {Psychology} and {Cognitive} {Neuroscience}}}, J.~T. Wixted, Ed.\hskip 1em plus 0.5em minus 0.4em\relax Hoboken, NJ, USA: John Wiley \& Sons, Inc., 2018.

\bibitem{Morris_Conversation1993O}
C.~W. Morris, ``On the importance of conversation,'' \emph{Dialogue}, vol.~32, 1993.

\bibitem{Bietti2010SharingMF}
L.~M. Bietti, ``Sharing memories, family conversation and interaction,'' \emph{Discourse \& Society}, vol.~21, 2010.

\bibitem{MemorabilityModeling2020}
A.~Newman, C.~Fosco, V.~Casser, A.~Lee, B.~McNamara, and A.~Oliva, ``Multimodal memorability: Modeling effects of semantics and decay on video memorability,'' in \emph{ECCV 2020}.\hskip 1em plus 0.5em minus 0.4em\relax Springer-Verlag, 2020, p. 223–240.

\bibitem{videoMemorability2018}
R.~Cohendet, K.~Yadati, N.~Q.~K. Duong, and C.-H. Demarty, ``Annotating, {{Understanding}}, and {{Predicting Long}}-term {{Video Memorability}},'' in \emph{Proceedings of the 2018 {{ACM}} on {{ICMR}} '18}, {New York, NY, USA}, 2018.

\bibitem{Isola2014ImageMemorability}
P.~Isola, J.~Xiao, D.~Parikh, A.~Torralba, and A.~Oliva, ``What makes a photograph memorable?'' \emph{IEEE IEEE Trans Pattern Anal Mach Intell}, vol.~36, 2014.

\bibitem{Bainbridge2017MemorabilityNeural}
W.~A. Bainbridge, D.~D. Dilks, and A.~Oliva, ``Memorability: A stimulus-driven perceptual neural signature distinctive from memory,'' \emph{NeuroImage}, vol. 149, 2017.

\bibitem{stafford_actor-observer_1989}
L.~Stafford, V.~R. Waldron, and L.~L. Infield, ``\BIBforeignlanguage{en}{Actor-{Observer} {Differences} in {Conversational} {Memory}},'' \emph{\BIBforeignlanguage{en}{Hum Comm Res}}, vol.~15, no.~4, 1989.

\bibitem{Heale2015Validity}
R.~Heale and A.~Twycross, ``Validity and reliability in quantitative studies,'' \emph{Evidence-Based Nursing}, vol.~18, 2015.

\bibitem{Rumpf2019EcolValMemory}
U.~Rumpf, I.~Menze, N.~G. M{\"u}ller, and M.~Schmicker, ``Investigating the potential role of ecological validity on change-detection memory tasks and distractor processing in younger and older adults,'' \emph{Front Psychol}, vol.~10, 2019.

\bibitem{Dunsmore2022SalientAttractMemory}
J.~E. Dunsmoor, V.~P. Murty, D.~Clewett, E.~A. Phelps, and L.~Davachi, ``Tag and capture: how salient experiences target and rescue nearby events in memory,'' \emph{Trends Cogn Sci}, vol.~26, no.~9, 2022.

\bibitem{Schnitzspahn2018InLabMemory}
K.~M. Schnitzspahn, L.~Kvavilashvili, and M.~Altgassen, ``Redefining the pattern of age-prospective memory-paradox: new insights on age effects in lab-based, naturalistic, and self-assigned tasks,'' \emph{Psychol Res}, vol.~84, 2018.

\bibitem{Jiang_Zhang_Choi_2020scripted}
H.~Jiang, X.~Zhang, and J.~D. Choi, ``\BIBforeignlanguage{en}{Automatic text-based personality recognition on monologues and multiparty dialogues using attentive networks and contextual embeddings},'' in \emph{\BIBforeignlanguage{en}{Proceedings of the CAI'20}}, vol.~34, no.~10.\hskip 1em plus 0.5em minus 0.4em\relax AAAI, 2020.

\bibitem{Daneman1986Individual}
M.~Daneman and I.~Green, ``Individual differences in comprehending and producing words in context,'' \emph{J Mem Lang}, vol.~25, 1986.

\bibitem{Erba2021CommunicationResearchDemographicsBias}
J.~Erba, P.~S. Bobkowski, B.~Ternes, Y.~Liu, and T.~Logan, ``Who are the “masses” in mass communication research? exploring participants’ demographic characteristics between 2000 and 2014,'' \emph{Howard J Commun}, vol.~33, 2021.

\bibitem{Infante-Rivard2018SelectionBias}
C.~Infante-Rivard and A.~Cusson, ``Reflection on modern methods: selection bias-a review of recent developments.'' \emph{Int J Epidemiol}, vol. 47 5, 2018.

\bibitem{Qureshi2022AmazonTurkDemographics}
N.~Qureshi, M.~Edelen, L.~Hilton, A.~Rodriguez, R.~D. Hays, and P.~Herman, ``Comparing data collected on amazon's mechanical turk to national surveys.'' \emph{Am J Health Behav}, vol. 46 5, 2022.

\bibitem{Cohendet2016MemorabilityByGender}
R.~Cohendet, A.-L. Gilet, M.~Perreira Da~Silva, and P.~Le~Callet, ``Using individual data to characterize emotional user experience and its memorability: Focus on gender factor,'' in \emph{2016 Eighth QoMEX}.\hskip 1em plus 0.5em minus 0.4em\relax IEEE, 2016.

\bibitem{Dudzik2018Lifelog}
B.~Dudzik, J.~Broekens, M.~Neerincx, J.~Olenick, C.-H. Chang, S.~W.~J. Kozlowski, and H.~Hung, ``\BIBforeignlanguage{en}{Discovering digital representations for remembered episodes from lifelog data},'' in \emph{\BIBforeignlanguage{en}{Proceedings of MCPMD '18}}.\hskip 1em plus 0.5em minus 0.4em\relax Boulder Colorado: ACM, Oct. 2018.

\bibitem{Cleary2018MemoryMeasures}
A.~M. Cleary, ``Dependent measures in memory research,'' \emph{Handbook of Research Methods in Human Memory}, 2018.

\bibitem{benoit_memory_1990}
W.~L. Benoit and P.~J. Benoit, ``\BIBforeignlanguage{en}{Memory for conversational behavior},'' \emph{\BIBforeignlanguage{en}{South Commun J}}, vol.~56, no.~1, 1990.

\bibitem{Patino_Ferreira_2018ValidityTradeoff}
C.~M. Patino and J.~C. Ferreira, ``\BIBforeignlanguage{en}{Internal and external validity: can you apply research study results to your patients?}'' \emph{\BIBforeignlanguage{en}{J Bras Pneumol}}, vol.~44, no.~3, 2018.

\bibitem{Jd1983MoodRecall}
T.~Jd and R.~Ml, ``Differential effects of induced mood on the recall of positive, negative and neutral words,'' \emph{Br J Clin Psychol}, vol.~22, 1983.

\bibitem{Matt1992Mood-congruent}
G.~Matt, C.~Vázquez, and W.~K. Campbell, ``Mood-congruent recall of affectively toned stimuli: A meta-analytic review,'' \emph{Clin Psychol Rev}, vol.~12, 1992.

\bibitem{Mayo1989PersonalityMemory}
P.~R. Mayo, ``A further study of the personality-congruent recall effect,'' \emph{Pers Individ Dif}, vol.~10, 1989.

\bibitem{Villaseñor2021ValuesMemory}
J.~J. Villaseñor, A.~Sklenar, A.~Frankenstein, P.~U. Levy, M.~P. McCurdy, and E.~Leshikar, ``Value-directed memory effects on item and context memory,'' \emph{Mem Cognit}, vol.~49, 2021.

\bibitem{IOS_woosnam2010inclusion}
K.~M. Woosnam, ``The inclusion of other in the self (ios) scale,'' \emph{Ann Tour Res}, vol.~37, no.~3, 2010.

\bibitem{Evans1991CohesionPerformance}
C.~R. Evans and K.~Dion, ``Group cohesion and performance,'' \emph{Small Group Res}, vol.~22, 1991.

\bibitem{Kim2020CohesionLearning}
S.-Y. Kim and E.~Yang, ``Does group cohesion foster self-directed learning for medical students? a longitudinal study,'' \emph{BMC Medical Education}, vol.~20, 2020.

\bibitem{cohendet_videomem_2019}
R.~Cohendet, C.-H. Demarty, N.~Duong, and M.~Engilberge, ``\BIBforeignlanguage{en}{{VideoMem}: {Constructing}, {Analyzing}, {Predicting} {Short}-{Term} and {Long}-{Term} {Video} {Memorability}},'' in \emph{\BIBforeignlanguage{en}{2019 {IEEE}/{CVF} ICCV)}}.\hskip 1em plus 0.5em minus 0.4em\relax Seoul, Korea (South): IEEE, 2019.

\bibitem{Bernd_mementos}
B.~Dudzik, H.~Hung, M.~Neerincx, and J.~Broekens, ``Collecting mementos: A multimodal dataset for context-sensitive modeling of affect and memory processing in responses to videos,'' \emph{IEEE TAC}, vol.~14, no.~2, 2023.

\bibitem{WoNoWA2020}
B.~Biancardi, L.~Maisonnave-Couterou, P.~Renault, B.~Ravenet, M.~Mancini, and G.~Varni, ``The wonowa dataset: Investigating the transactive memory system in small group interactions,'' in \emph{Proceedings of the ICMI'20}.\hskip 1em plus 0.5em minus 0.4em\relax New York, NY, USA: ACM, 2020.

\bibitem{Shaw2007homebirth}
R.~Shaw and C.~Kitzinger, ``Memory in interaction: an analysis of repeat calls to a home birth helpline,'' \emph{Res Lang Soc Interact}, vol.~40, 2007.

\bibitem{Tulving1966AvailabileAccessibleMem}
E.~Tulving and Z.~Pearlstone, ``Availability versus accessibility of information in memory for words,'' \emph{J Verb Learn Verb Behav}, vol.~5, 1966.

\bibitem{Murre2015ForgettingCurve}
J.~M.~J. Murre and J.~Dros, ``Replication and analysis of ebbinghaus’ forgetting curve,'' \emph{PLOS ONE}, vol.~10, no.~7, 2015.

\bibitem{2022MedievalEEGMem}
A.~García Seco~de Herrera, M.~G. Constantin, C.-H. Demarty, C.~Fosco, S.~Halder, G.~Healy, B.~Ionescu, A.~Matran-Fernandez, A.~F. Smeaton, M.~Sultana, and L.~Sweeney, ``\BIBforeignlanguage{en}{Experiences from the mediaeval predicting media memorability task},'' in \emph{\BIBforeignlanguage{en}{The NeurIPS MemARI Workshop proceedings}}, New Orleans, USA, 2022.

\bibitem{prolific}
\BIBentryALTinterwordspacing
``Prolific · quickly find research participants you can trust.'' 2022. [Online]. Available: \url{http://www.prolific.co/}
\BIBentrySTDinterwordspacing

\bibitem{Wheelan2009GroupSize}
S.~Wheelan, ``Group size, group development, and group productivity,'' \emph{Small Group Res}, vol.~40, 2009.

\bibitem{Hexaco_DEVRIES2013871}
R.~E. {de Vries}, ``The 24-item brief hexaco inventory (bhi),'' \emph{J Res Pers}, vol.~47, no.~6, 2013.

\bibitem{Schwartz_vals_2005}
M.~Lindeman and M.~Verkasalo, ``Measuring values with the short schwartz's value survey,'' \emph{J Pers Assess}, vol.~85, no.~2, 2005.

\bibitem{qualtrics}
\BIBentryALTinterwordspacing
Oct 2021. [Online]. Available: \url{https://www.qualtrics.com/}
\BIBentrySTDinterwordspacing

\bibitem{broekens2013affectbutton}
J.~Broekens and W.-P. Brinkman, ``Affectbutton: A method for reliable and valid affective self-report,'' \emph{Int J Hum Comput Stud}, vol.~71, no.~6, 2013.

\bibitem{aron1992IOS}
A.~Aron, E.~N. Aron, and D.~Smollan, ``Inclusion of other in the self scale and the structure of interpersonal closeness.'' \emph{J Pers Soc Psychol}, vol.~63, no.~4, 1992.

\bibitem{braun2020cohesion}
M.~T. Braun, S.~W. Kozlowski, T.~A. Brown, and R.~P. DeShon, ``Exploring the dynamic team cohesion--performance and coordination--performance relationships of newly formed teams,'' \emph{Small Group Res}, vol.~51, no.~5, 2020.

\bibitem{koudenburg2014entitativity}
N.~Koudenburg, T.~Postmes, and E.~H. Gordijn, ``Conversational flow and entitativity: The role of status,'' \emph{Br J Clin Psychol}, vol.~53, no.~2, 2014.

\bibitem{gerpott2018interdependence}
F.~H. Gerpott, D.~Balliet, S.~Columbus, C.~Molho, and R.~E. de~Vries, ``How do people think about interdependence? a multidimensional model of subjective outcome interdependence.'' \emph{J Pers Soc Psychol}, vol. 115, no.~4, 2018.

\bibitem{DIAMONDSrauthmann2016ultra}
J.~F. Rauthmann and R.~A. Sherman, ``Ultra-brief measures for the situational eight diamonds domains.'' \emph{Eur J Psychol Assess}, vol.~32, no.~2, 2016.

\bibitem{Kurby2008MemorySegmentation}
C.~A. Kurby and J.~M. Zacks, ``\BIBforeignlanguage{en}{Segmentation in the perception and memory of events},'' \emph{\BIBforeignlanguage{en}{Trends Cogn Sci}}, vol.~12, no.~2, 2008.

\bibitem{Zoom}
\BIBentryALTinterwordspacing
``\BIBforeignlanguage{en-US}{Zoom video conferencing platform}.'' [Online]. Available: \url{https://zoom.us}
\BIBentrySTDinterwordspacing

\bibitem{Povey_ASRU2011}
D.~e.~a. Povey, ``The kaldi speech recognition toolkit,'' in \emph{IEEE 2011 Workshop on Automatic Speech Recognition and Understanding}.\hskip 1em plus 0.5em minus 0.4em\relax IEEE SPS, 2011.

\bibitem{eyeware_2022}
\BIBentryALTinterwordspacing
GazeSense, ``3d eye tracking software,'' Jul 2022. [Online]. Available: \url{https://eyeware.tech/gazesense/}
\BIBentrySTDinterwordspacing

\bibitem{amos_openface_2016}
B.~Amos, B.~Ludwiczuk, and M.~Satyanarayanan, ``{OpenFace}: {A} general-purpose face recognition library with mobile applications,'' CMU-CS-16-118, CMU School of Computer Science, Tech. Rep., 2016.

\bibitem{Lugaresi2019MediaPipeAF}
C.~Lugaresi, J.~Tang, H.~Nash, C.~McClanahan, E.~Uboweja, M.~Hays, F.~Zhang, C.-L. Chang, M.~G. Yong, J.~Lee, W.-T. Chang, W.~Hua, M.~Georg, and M.~Grundmann, ``Mediapipe: A framework for building perception pipelines,'' \emph{ArXiv}, vol. abs/1906.08172, 2019.

\bibitem{baltruvsaitis2016openface}
T.~Baltru{\v{s}}aitis, P.~Robinson, and L.-P. Morency, ``Openface: an open source facial behavior analysis toolkit,'' in \emph{2016 IEEE WACV}.\hskip 1em plus 0.5em minus 0.4em\relax IEEE, 2016.

\bibitem{Cohen1982Primacy}
R.~L. Cohen, ``On the generality of some memory laws,'' \emph{Scand J Psychol}, vol.~22, no.~1, 1981.

\bibitem{Tsfasman2022}
M.~Tsfasman, K.~Fenech, M.~Tarvirdians, A.~Lorincz, C.~Jonker, and C.~Oertel, ``\BIBforeignlanguage{en}{Towards creating a conversational memory for long-term meeting support: predicting memorable moments in multi-party conversations through eye-gaze},'' in \emph{\BIBforeignlanguage{en}{Proceedings of the 2022 ICMI}}.\hskip 1em plus 0.5em minus 0.4em\relax Bengaluru India: ACM, 2022.

\bibitem{Gächter2015IOSClosenes}
S.~Gächter, C.~Starmer, and F.~Tufano, ``Measuring the closeness of relationships: A comprehensive evaluation of the 'inclusion of the other in the self' scale,'' \emph{PLoS ONE}, vol.~10, 2015.

\bibitem{Clark2019CommonGroundNeedCA}
L.~Clark, N.~Pantidi, O.~Cooney, P.~R. Doyle, D.~Garaialde, J.~Edwards, B.~Spillane, C.~Murad, C.~Munteanu, V.~Wade, and B.~R. Cowan, ``What makes a good conversation?: Challenges in designing truly conversational agents,'' \emph{Proceedings of the 2019 CHI}, 2019.

\bibitem{Wrede2003SpottingS}
B.~Wrede and E.~Shriberg, ``Spotting "hot spots" in meetings: human judgments and prosodic cues,'' in \emph{INTERSPEECH}, 2003.

\bibitem{Nihei2019SummarisationImportance}
F.~Nihei and Y.~I. Nakano, ``Exploring methods for predicting important utterances contributing to meeting summarization,'' \emph{Multimodal Technol Interact}, vol.~3, no.~3, 2019.

\bibitem{Akhter-Khan2020Why}
S.~C. Akhter-Khan and R.~Au, ``Why loneliness interventions are unsuccessful: A call for precision health,'' \emph{Adv Geriatr Med Res}, vol.~2, 2020.

\bibitem{Mitsuki}
E.~Croes and M.~L. Antheunis, ``Can we be friends with mitsuku? a longitudinal study on the process of relationship formation between humans and a social chatbot,'' \emph{J Soc Pers Relat}, vol.~38, 2020.

\bibitem{elvir_remembering_2017}
M.~Elvir, A.~J. Gonzalez, C.~Walls, and B.~Wilder, ``\BIBforeignlanguage{en}{Remembering a {Conversation} – {A} {Conversational} {Memory} {Architecture} for {Embodied} {Conversational} {Agents}},'' \emph{\BIBforeignlanguage{en}{J Intell Syst}}, vol.~26, no.~1, 2017.

\bibitem{brom_episodic_2010}
C.~Brom, J.~Lukavský, and R.~Kadlec, ``Episodic memory for human-like agents and human-like agents for episodic memory,'' \emph{Int J Mach Consc}, vol.~02, no.~02, 2010.

\bibitem{kasap_building_2012}
Z.~Kasap and N.~Magnenat-Thalmann, ``\BIBforeignlanguage{en}{Building long-term relationships with virtual and robotic characters: the role of remembering},'' \emph{\BIBforeignlanguage{en}{The Visual Computer}}, vol.~28, no.~1, 2012.

\bibitem{Nash2022}
A.~Saravanan, M.~Tsfasman, M.~A. Neerincx, and C.~Oertel, ``Giving social robots a conversational memory for motivational experience sharing,'' in \emph{2022 31st IEEE RO-MAN}, 2022.

\bibitem{Tian2023TurntakingOnlineVSOffline}
Y.~Tian, S.~Liu, and J.~Wang, ``A corpus study on the difference of turn-taking in online audio, online video, and face-to-face conversation.'' \emph{Lang Speech}, 2023.

\bibitem{Ng2007OnlineRecallEqualsOffline}
C.~S.~L. Ng and W.~Cheung, ``Comparing face to face, tutor led discussion and online discussion in the classroom,'' \emph{Aust J Ed Techn}, vol.~23, 2007.

\bibitem{Ray2019OMemoryImportance}
D.~G. Ray, S.~Gomillion, A.~I. Pintea, and I.~Hamlin, ``On being forgotten: Memory and forgetting serve as signals of interpersonal importance,'' \emph{J Pers Soc Psychol}, vol. 116, 2019.

\bibitem{FinalCutPro}
\BIBentryALTinterwordspacing
``\BIBforeignlanguage{nl-NL}{Final cut pro}.'' [Online]. Available: \url{https://www.apple.com/nl/final-cut-pro/}
\BIBentrySTDinterwordspacing

\end{thebibliography}

\section*{Author Biographies}

\begin{IEEEbiography}
[{\includegraphics[width=1in,height=1.75in,clip,keepaspectratio]{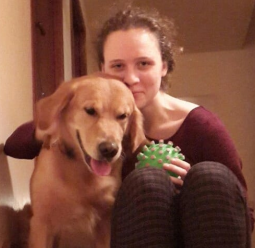}}]{Maria Tsfasman} (a.k.a Masha) Her interdisciplinary memory research aims to enhance machine understanding of humans and gain insights into human cognition. She holds a BSc in Linguistics from HSE University, Moscow, and a MSc in AI with distinction from Radboud University, Nijmegen. Before her PhD, she is proud to have been a research assistant at ISIR, Sorbonne University, and IRCN, University of Tokyo. 
\end{IEEEbiography}

\begin{IEEEbiography}
[{\includegraphics[width=1in,height=1.75in,clip,keepaspectratio]{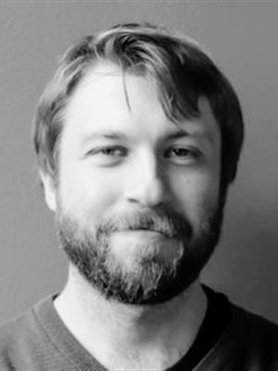}}]{Bernd Dudzik} is an Assistant Professor in the Department of Intelligent Systems at \textit{TU Delft}. His research interests focus on Affective Computing and AI-Human collaboration. In particular, his work explores context-sensitive approaches for multimodal modeling of human cognitive-affective processes (e.g., memory recollection or cognitive appraisals) during everyday interactions. Bernd is an active member of the AAAC and Associate Editor for the IEEE Transactions on Affective Computing. \end{IEEEbiography}

\begin{IEEEbiography}
[{\includegraphics[width=1in,height=1.75in,clip,keepaspectratio]{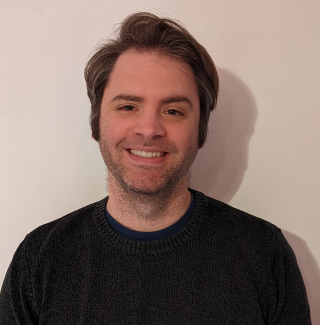}}]{Kristian Fenech}  received a Ph.D. degree in Physics from Swinburne University of Technology, Melbourne, Australia, in 2016. He is currently an assistant professor with the Department of Artificial Intelligence, Faculty of Informatics, Eötvös Loránd University, Budapest, Hungary. His research interests include human centered AI and human-machine interaction.
\end{IEEEbiography}

\begin{IEEEbiography}
[{\includegraphics[width=1in,height=1.75in,clip,keepaspectratio]{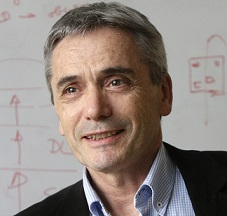}}]{András Lőrincz}'s research lies at the interface of artificial general intelligence, cognition and neuroscience, with his main areas of interest being intelligent systems, human-computer interaction and cooperation, social cognition, modelling and applications. He is a Fellow of the European Association for Artificial Intelligence (EurAI) and a Member of the European Laboratory for Learning and Intelligent Systems (ELLIS). His background is in physics, received his PhD in chemical physics. 
\end{IEEEbiography}

\begin{IEEEbiography}
[{\includegraphics[width=1in,height=1.75in,clip,keepaspectratio]{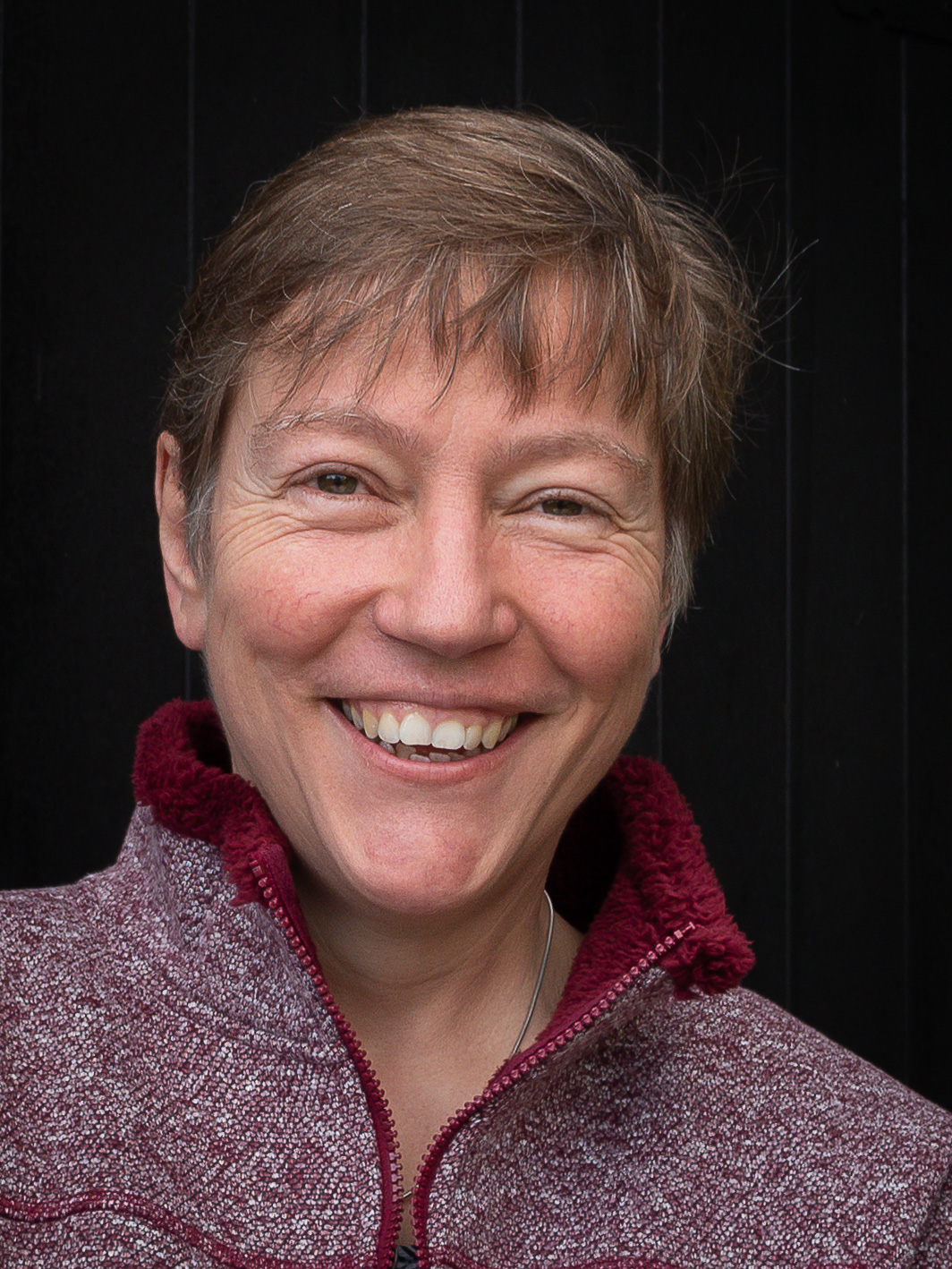}}]{Catholijn M. Jonker} researches hybrid intelligence and interactive intelligent processes such as strategic decision making, negotiation, teamwork and the design of agent technology. She is board member of IFAAMAS, and vice-coordinator of the Hybrid Intelligence Centre. She is a member of the Royal Holland Society of Sciences and Humanities, a Fellow of EurAI, and member of the Academia Europaea. She is a co-founding member of the Netherlands Academy of Engineers. 
\end{IEEEbiography}

\begin{IEEEbiography}
[{\includegraphics[width=1in,height=1.75in,clip,keepaspectratio]{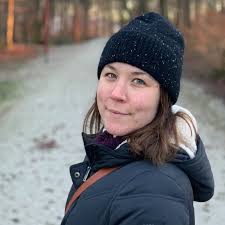}}]{Catharine Oertel} received her PhD degree from the Royal Institute of Technology (KTH), Sweden. After completing her postdoctoral position at EPFL, Switzerland she joined TU Delft as an assistant professor in 2019. Her research focuses on long-term human agent interaction where she particularly focuses on memory modelling.
\end{IEEEbiography}

%will be removed for submission
% \documentclass[lettersize, journal, compsoc]{IEEEtran}
% \usepackage{amsmath,amsfonts}
% \usepackage{algorithmic}
% \usepackage{algorithm}
% \usepackage{array}
% \usepackage[caption=false,font=normalsize,labelfont=sf,textfont=sf]{subfig}
% \usepackage{textcomp}
% \usepackage{stfloats}
% \usepackage{url}
% \usepackage{verbatim}
% \usepackage{graphicx}
% \usepackage{cite}
% %\hyphenation{op-tical net-works semi-conduc-tor IEEE-Xplore}
% \usepackage{multirow}
% \usepackage{pifont}
% %\usepackage[noadjust]{cite}  
% \usepackage[inline]{enumitem}
% % updated with editorial comments 8/9/2021
% \usepackage[textsize=tiny]{todonotes}
% \usepackage{hyperref}
% \usepackage{float}

% \begin{document}
\onecolumn
\setcounter{section}{0}
\setcounter{table}{0}
\section*{Appendix}
\section{Questionnaires used in MEMO corpus \label{appendix:surveys}}
Table \ref{tab:questionnaires} below shows a full list of measures collected within the surveys in MEMO data collection. See detailed descriptions of each measure in \emph{Section \ref{procedure}}.

\begin{table*}[!ht]
\centering
\caption{A full list of measures collected within the prescreening, pre-session and post-session surveys. Note that the order of the mentioned measures does not correspond to the order in which the measures were presented to participants. '-' in scale column means that the measures were 1-item questions rather than a full validated scale (see table below for specific questions). }
\label{tab:questionnaires}
\begin{tabular}{l|l|l|l|l}
\hline \hline
\textbf{survey} & \textbf{category} & \textbf{variable} & \textbf{scale} & \textbf{when} \\ \hline
\multirow{6}{*}{\textbf{\begin{tabular}[c]{@{}l@{}}prescreening\\ survey\end{tabular}}} & demographics & \begin{tabular}[c]{@{}l@{}}Age, gender, English fluency, \\ country of residency, \\ COVID-19 affected group\end{tabular} & - & \multirow{6}{*}{\begin{tabular}[c]{@{}l@{}}before\\ the\\ experiment\end{tabular}} \\ \cline{2-4}
 & \multirow{3}{*}{\begin{tabular}[c]{@{}l@{}}personal \\ charachteristics\end{tabular}} & Personality & \begin{tabular}[c]{@{}l@{}}24-item Brief \\ HEXACO Inventory \cite{Hexaco_DEVRIES2013871}\end{tabular} &  \\ \cline{3-4}
 &  & Values & \begin{tabular}[c]{@{}l@{}}The Short Schwartz’s\\ Value Survey \cite{Schwartz_vals_2005}\end{tabular} &  \\ \cline{3-4}
 &  & \begin{tabular}[c]{@{}l@{}}Experience with online \\ meetings\end{tabular} & - &  \\ \cline{2-4}
 & consent & \begin{tabular}[c]{@{}l@{}}Consent for recording and \\ storing video, audio and \\ survey data\end{tabular} & - &  \\ \cline{2-4}
 & \begin{tabular}[c]{@{}l@{}}technical requirements\\ verification\end{tabular} & \begin{tabular}[c]{@{}l@{}}Laptop with working camera\\ and headset with a working \\ microphone\end{tabular} & - &  \\ \hline
\multirow{4}{*}{\textbf{\begin{tabular}[c]{@{}l@{}}pre-session\\ survey\end{tabular}}} & \begin{tabular}[c]{@{}l@{}}mood at the start of\\ the session\end{tabular} & Mood before the session & Affect Button \cite{broekens2013affectbutton} & \multirow{2}{*}{\begin{tabular}[c]{@{}l@{}}right before\\ each conversation\\ session\end{tabular}} \\ \cline{2-4}
 & screenshot upload & \begin{tabular}[c]{@{}l@{}}Zoom setup for gaze target\\ extraction\end{tabular} & - &  \\ \cline{2-5} 
 & \begin{tabular}[c]{@{}l@{}}memory of the previous\\ session\end{tabular} & Free recall & - & \begin{tabular}[c]{@{}l@{}}before 2nd \&\\ 3rd sessions \&\\ exit\\ interview\end{tabular} \\ \cline{2-5} 
 & \begin{tabular}[c]{@{}l@{}}most important moment\\ of previous discussion \\ (start and end times)\end{tabular} & \begin{tabular}[c]{@{}l@{}}Most important moment \\ for grounding the questions\\ in the exit interview\end{tabular} & - & \begin{tabular}[c]{@{}l@{}}before \\ exit\\ interview\end{tabular} \\ \hline
\multirow{15}{*}{\textbf{\begin{tabular}[c]{@{}l@{}}post-session\\ survey\end{tabular}}} & \multirow{3}{*}{memory} & Free recall & - & \multirow{15}{*}{\begin{tabular}[c]{@{}l@{}}straight after\\ each conversation\\ session\end{tabular}} \\ \cline{3-4}
 &  & Timing annotation & - &  \\ \cline{3-4}
 &  & Reason for remembering & - &  \\ \cline{2-4}
 & memory for conv. agent & - & - &  \\ \cline{2-4}
 & \multirow{6}{*}{\begin{tabular}[c]{@{}l@{}}perception of\\ the group and interaction\end{tabular}} & Task \& Group Cohesion & \begin{tabular}[c]{@{}l@{}}Cohesion in newly \\ formed teams \cite{braun2020cohesion}\end{tabular} &  \\ \cline{3-4}
 &  & Entitativity & Entitativity Scale \cite{koudenburg2014entitativity} &  \\ \cline{3-4}
 &  & Syncness & - &  \\ \cline{3-4}
 &  & Rapport & - &  \\ \cline{3-4}
 &  & Perceived Interdependence & \begin{tabular}[c]{@{}l@{}}Situational\\ Interdependence \cite{gerpott2018interdependence}\end{tabular} &  \\ \cline{3-4}
 &  & \begin{tabular}[c]{@{}l@{}}Perceived Situation\\ Characteristics\end{tabular} & DIAMONDS \cite{DIAMONDSrauthmann2016ultra} &  \\ \cline{2-4}
 & \multirow{5}{*}{\begin{tabular}[c]{@{}l@{}}perception of \\ other participants\\ (one by one)\end{tabular}} & Relationship (perceived distance) & IOS scale \cite{IOS_woosnam2010inclusion} &  \\ \cline{3-4}
 &  & \begin{tabular}[c]{@{}l@{}}Mutual understanding \\ (perceived values)\end{tabular} & \begin{tabular}[c]{@{}l@{}}The Short Schwartz’s\\ Value Survey \cite{Schwartz_vals_2005}\end{tabular} &  \\ \cline{3-4}
 &  & Quality as a listener & - &  \\ \cline{3-4}
 &  & Personal attitude & - &  \\ \cline{3-4}
 &  & \begin{tabular}[c]{@{}l@{}}Quality as conversational\\ partner\end{tabular} & - &  \\ \hline \hline
\end{tabular}

\end{table*}

\newpage
\section{Formulations of original questionnaire}
Table \ref{tab:survey_extra_items} lists the specific formulations of the original questionnaire items that were created specifically for this dataset and were not directly adopted from existing validated scales.

\begin{table*}[!htbp]
\centering
\caption{A list of question formulations for the measures that were specific to this dataset (the ones that were not adopted from an existing validated scale)\label{tab:survey_extra_items}}

\begin{tabular}{l|l|c|c}
\hline
\multicolumn{1}{c|}{\textbf{Measure}} & \multicolumn{1}{c|}{\textbf{Question formulation}} & \textbf{\begin{tabular}[c]{@{}c@{}}Type \\ of response\end{tabular}} & \textbf{\begin{tabular}[c]{@{}c@{}}Which\\ questionnaire\end{tabular}} \\ \hline
\textbf{\begin{tabular}[c]{@{}l@{}}Free recall\\ reports\end{tabular}} & \begin{tabular}[c]{@{}l@{}}"Recall and describe moments of the most recent\\  discussion session in as much detail as you can\\  remember. Any details are great - for example, \\ about the content, other participants, the moderator, \\ you, your feelings, the reaction of others, your words, \\ others' words, timing, or anything that happened \\ throughout the discussion. Recall at least 3 moments. \\ If you remember more, the fields will show up as \\ you go until you leave one of them empty.\\
...\\
Write the [\textit{first/ second/ third}] moment you remember\\ from the last discussion that you had. The more\\ details
the better :)\\
...
\\
Do you remember another moment from the last\\discussion that you had? In case you remember\\ more moments, another field will show up on\\the next page. If you don't remember more moments \\leave the box empty and proceed.
"\end{tabular} & text & \begin{tabular}[c]{@{}c@{}}Post-questionnaire \\ (after all sessions)\\ and \\ Pre-questionnaire\\ (before 2nd, 3rd\\ and exit interview)\end{tabular} \\ \hline
\textbf{\begin{tabular}[c]{@{}l@{}}Timing\\ annotation\end{tabular}} & \begin{tabular}[c]{@{}l@{}}"You wrote down several moments you remembered\\ from the previous discussion at the beginning of the\\ survey. Can you now open the video recording and\\ try to find when those moments occurred? It's ok if\\ the timestamp is not too precise. Please, don't rewatch\\ the whole video, only use it to look up the specific time\\ of each memory moment you wrote down. You can\\ move your cursor along the timeline of the video to go\\ to a specific moment. Please close it as soon as you are\\ done with this survey and don't come back to it until \\ you finish the experiment (officially finished the entire\\ experiment on prolific). If you can't find the exact \\ moment, put "0" in the time fields and fill in the option \\ 'Comment' with any details you remember of the timing.\\ Moment{[}N\_\{memory\}{]}: At what point in the conversation \\ this moment happened: "{[}quote from the free recall\\ report {[}N\_\{memory\}{]}"."\end{tabular} & numbers & \multirow{6}{*}{Post-questionnaire} \\ \cline{1-3}
\textbf{\begin{tabular}[c]{@{}l@{}}Reason for \\ remembering\end{tabular}} & \begin{tabular}[c]{@{}l@{}}"Write down why do you think this moment was \\ memorable for you."\end{tabular} & text &  \\ \cline{1-3}
\textbf{\begin{tabular}[c]{@{}l@{}}Memory for \\ conversational \\ agent question\end{tabular}} & \begin{tabular}[c]{@{}l@{}}"We want to build a social robot that really \\ understands you and what is important to you\\  and represent you in the future meetings. It\\  could represent you in the discussions with\\  other people to make sure your perspective is\\  being heard. It could also serve as your \\ personal brain-storming partner with whom \\ you can deliberate important aspects of your \\ life or decisions you need to take. What would\\  be important for such a robot to remember from \\ this meeting? Which specific moments of the\\ conversation would you want it to remember? \\ What details are most important to remember? \\ Write as many details as you can."\end{tabular} & text &  \\ \cline{1-3}
\textbf{\begin{tabular}[c]{@{}l@{}}Quality as a\\ listener\end{tabular}} & \begin{tabular}[c]{@{}l@{}}"To what extent do other participants have the\\ following qualities? \\ - To what extent is {[}N\_\{participant\}{]} a good listener? "\end{tabular} & \multirow{3}{*}{\begin{tabular}[c]{@{}c@{}}7-point likert scale\\  (1=not at all to\\  7=very much)\end{tabular}} &  \\ \cline{1-2}
\textbf{\begin{tabular}[c]{@{}l@{}}Personal \\ attitude\end{tabular}} & "- To what extent do you like {[}N\_\{participant\}{]}?" &  &  \\ \cline{1-2}
\textbf{\begin{tabular}[c]{@{}l@{}}Quality as \\ conversational\\ partner\end{tabular}} & \begin{tabular}[c]{@{}l@{}}"- How would you rate {[}N\_\{participant\}{]}’s ability to\\ keep the conversation flowing"\end{tabular} &  &  \\ \hline
\end{tabular}
\end{table*}

\newpage

\section{Additional recruitment criteria}
\label{appendix:recr_criteria}
Following are additional recruitment criteria per target group we used to diversify opinions in each group:
\begin{itemize}
    \item 'Parents with young children': having at least one 2-12 y.o. child at the time of the pandemic.
    \item 'Students': an active student status.
    \item 'Older adults': 50 years of age or older (since it was the COVID-19 risk group)
    \item '(Ex-)business owners': having had an entrepreneur status in the last 4 years before the experiment.
\end{itemize} 

Using these criteria, the resulting sample included 9 parents of young children, 13 students,  10 business owners, 14 older adults (50+), and 7 others, not identifying as parts of the groups above. 

\section{Separated audio synchronisation procedure}
\label{synchronisation}
Because of a technical issue with the used version of Zoom, these audios were not synchronised to the video. We have synchronised these audios to the original video using Final Cut’s synchronise feature \cite{FinalCutPro}. The audio was then manually checked for inconsistencies and corrected to match the original audio. We then plotted the sum of all separated audios against the original audio and calculated the absolute difference between the two audios to find any missed non-synchronised segments. The resulting synchronised separated audio channels are recorded in the original .m4a format, with separate audio for each participant in every session. Since participants used their own computers and headsets, the audio might vary in quality, reflecting an ecologically valid setting of a video-call set-up (see P1.1 in \emph{Section \ref{principles}}).

\section{Longitudinal questionnaire completeness}
\emph{Table \ref{appendix:completeness}} shows the percentage of participants (including the moderator) that completed pre-and/ or post-questionnaire over all 3 sessions.
\begin{table}[!ht]

    \centering
\caption{Questionnaire completeness index per group (the percentage of participants that completed pre- and/or post-survey over all the sessions) \label{appendix:completeness}}
\begin{tabular}{cccc}
\hline \hline
group & \multicolumn{3}{r}{\textbf{questionnaire completeness}} \\
number & \textbf{pre-survey} & \textbf{post-survey} & \textbf{total} \\
\hline

\textbf{1} & 0.80 & 0.80 & 0.80 \\
\textbf{2} & 0.80 & 0.60 & 0.60 \\
\textbf{3} & 0.50 & 0.75 & 0.50 \\
\textbf{4} & 0.83 & 0.67 & 0.67 \\
\textbf{5} & 1.00 & 1.00 & 1.00 \\
\textbf{6} & 0.60 & 0.80 & 0.40 \\
\textbf{7} & 0.75 & 1.00 & 0.75 \\
\textbf{8} & 0.75 & 0.75 & 0.50 \\
\textbf{9} & 0.50 & 1.00 & 0.50 \\
\textbf{10} & 0.83 & 0.83 & 0.83 \\
\textbf{11} & 1.00 & 1.00 & 1.00 \\
\textbf{12} & 1.00 & 0.75 & 0.75 \\
\textbf{13} & 0.75 & 1.00 & 0.75 \\
\textbf{14} & 0.50 & 0.25 & 0.25 \\
\textbf{15} & 1.00 & 1.00 & 1.00 \\
\hline \hline
\end{tabular}

\end{table}

\section{Dataset version statistics}
\emph{Table \ref{curationstats}} shows the statistics of the data included in each version of the dataset: pseudo-anonymised, processed and curated.
\begin{table}[H]
    \centering
    
\caption{Statistics of the data included in each publicly available version of the dataset\label{curationstats}}
\begin{tabular}{lccc}
\hline \hline
 & \textbf{Pseudo-anonymised} & \textbf{Processed} & \textbf{Curated} \\
 &  &  &  \\
\hline

\textbf{$N_{groups}$} & 15 & 15 & 15 \\
\textbf{$N_{sessions}$} & 45 & 45 & 44 \\
\textbf{$N_p$ in conversation} & 55 & 55 & 54 \\
\textbf{$N_p$ in pre-screening survey} & 154 & 55 & 54 \\
\textbf{$N_p$ in post-questionnaire} & 54 & 54 & 53 \\
\textbf{$N_p$ in pre-questionnaire} & 54 & 54 & 53 \\
\textbf{recording duration (min)} & 2050 & 1942 & 1892 \\
\textbf{recording duration (hours)} & 34 & 32 & 31 \\
\textbf{memorable segments count} & 853 & 694 & 622 \\
\textbf{memory duration in min (M} & 3.27 & 3.27 & 1.88 \\
\textbf{ +- STD)} & +- 6.6 & +- 6.6 & +- 2.01 \\
\hline \hline
\end{tabular}

\end{table}
\vfill

\section{Questionnaire descriptive statistics}
\emph{Table \ref{appendix:queststats}}  shows mood and situation perception scores across the curated version of the MEMO dataset.
\begin{table}[!htbp]

    \centering
\caption{Descriptive statistics of mood and DIAMONDS situation perception across MEMO data \label{appendix:queststats}}

\begin{tabular}{ccccc}
\hline \hline
 & \textbf{mean} & \textbf{std} & \textbf{min} & \textbf{max} \\

\hline
\textbf{AffectButton: Pleasure} & 0.35 & 0.38 & -0.61 & 1.00 \\
\textbf{AffectButton : Arousal} & -0.14 & 0.77 & -1.00 & 1.00 \\
\textbf{AffectButton : Dominance} & 0.22 & 0.57 & -1.00 & 1.00 \\
\textbf{DIAMONDS: Duty} & 3.88 & 1.62 & 1.00 & 7.00 \\
\textbf{DIAMONDS: Intellect} & 4.74 & 1.44 & 1.00 & 7.00 \\
\textbf{DIAMONDS: Adversity} & 1.42 & 0.91 & 1.00 & 6.00 \\
\textbf{DIAMONDS: Mating} & 1.26 & 0.78 & 1.00 & 5.00 \\
\textbf{DIAMONDS: Positivity} & 5.75 & 1.03 & 3.00 & 7.00 \\
\textbf{DIAMONDS: Negativity} & 2.28 & 1.41 & 1.00 & 7.00 \\
\textbf{DIAMONDS: Deception} & 1.61 & 1.17 & 1.00 & 6.00 \\
\textbf{DIAMONDS: Sociality} & 5.19 & 1.51 & 1.00 & 7.00 \\
\hline \hline
\end{tabular}

\end{table}
% \bibliographystyle{IEEEtran}  
% \bibliography{bib/short_references.bib}  
% \end{document}

\end{document}